\documentclass[10pt,twocolumn,letterpaper]{article}

\usepackage[pagenumbers]{cvpr}
\usepackage[table]{xcolor}

\newcommand{\red}[1]{{\color{red}#1}}
\newcommand{\todo}[1]{{\color{red}#1}}

\usepackage{amsmath,amsfonts,bm}

\def\eqref#1{equation~\ref{#1}}

\def\1{\bm{1}}

\def\rvz{{\mathbf{z}}}

\DeclareMathAlphabet{\mathsfit}{\encodingdefault}{\sfdefault}{m}{sl}
\SetMathAlphabet{\mathsfit}{bold}{\encodingdefault}{\sfdefault}{bx}{n}

\def\gD{{\mathcal{D}}}

\def\gP{{\mathcal{P}}}

\newcommand{\reg}{\lambda}

\definecolor{ForestGreen}{RGB}{34,139,34}

\definecolor{DarkenedMagenta}{RGB}{225,0,167}

\newcommand{\cls}{\textsc{[cls]}\xspace}

\newcommand{\reglue}{\texttt{REGLUE}\xspace}
\newcommand{\repae}{\texttt{REPA-E}\xspace}
\newcommand{\redi}{\texttt{ReDi}\xspace}
\newcommand{\repa}{\texttt{REPA}\xspace}
\renewcommand{\reg}{\texttt{REG}\xspace}
\newcommand{\rae}{\texttt{RAE}\xspace}
\newcommand{\vavae}{\texttt{VA-VAE}\xspace}
\newcommand{\sdvae}{\texttt{SD-VAE}\xspace}
\newcommand{\sdvaeftema}{\texttt{SD-VAE-FT-EMA}\xspace}
\newcommand{\dinovtwob}{\texttt{DINOv2-B}\xspace}
\newcommand{\dinovtwo}{\texttt{DINOv2}\xspace}
\newcommand{\imagenet}{ImageNet-1k\xspace}

\newcommand{\vitl}{ViT-L\xspace}
\newcommand{\vitb}{ViT-B\xspace}

\newcommand{\sit}{\texttt{SiT}\xspace}
\newcommand{\sitxl}{\texttt{SiT-XL/2}\xspace}
\newcommand{\repaxl}{\texttt{REPA-XL/2}\xspace}

\newcommand{\regluexl}{\texttt{REGLUE-XL/2}\xspace}
\newcommand{\regluebase}{\texttt{REGLUE-B/2}\xspace}
\newcommand{\regxl}{\texttt{REG-XL/2}\xspace}
\newcommand{\regbase}{\texttt{REG-B/2}\xspace}
\newcommand{\redixl}{\texttt{ReDi-XL/2}\xspace}
\newcommand{\coredi}{\texttt{CoReDi}\xspace}
\newcommand{\coredixl}{\texttt{CoReDi-XL/2}\xspace}
\newcommand{\rfid}{rFID\xspace}
\newcommand{\gfid}{gFID\xspace}

\newcommand{\hl}{\cellcolor{spteal!15}}
\newcommand{\redx}{\textcolor{red}{\ding{55}}}
\definecolor{uncomp}{RGB}{226,236,250}
\newcommand{\uc}{\rowcolor{uncomp}\cellcolor{white}}

\definecolor{Orange}{RGB}{255, 125, 0}
\definecolor{Purple}{RGB}{0.58, 0.44, 0.86}
\definecolor{VibrantMagenta}{RGB}{255,0,197}
\definecolor{TableColorAdditional}{rgb}{0.882, 0.945, 0.882}
\definecolor{TableColorExtra}{rgb}{0.906, 0.855, 0.941}
\definecolor{TableColor}{rgb}{0.80, 0.90, 0.93}
\definecolor{TableColorGrey}{rgb}{0.96, 0.96, 0.96}

\definecolor{TableColorGreyText}{rgb}{0.55, 0.55, 0.55}          
\definecolor{TableColorExtraText}{rgb}{0.60, 0.45, 0.80}         
\definecolor{TableColorAdditionalText}{rgb}{0.45, 0.65, 0.45}    
\definecolor{TableColorText}{rgb}{0.25, 0.60, 0.70}              
\definecolor{RegRowText}{rgb}{0.78, 0.68, 0.25}                  

\newcommand{\parag}[1]{\smallskip\noindent\textbf{#1}\enspace}
\newcommand{\paragnoskip}[1]{\noindent\textbf{#1}\enspace}

\def\addlegendimage{\csname pgfplots@addlegendimage\endcsname}

\newcommand{\supplautoref}[1]{\hyperref[#1]{Suppl.~\autoref{#1}}}
\newcommand{\supplsec}[1]{\hyperref[#1]{Suppl.~Sec.~\ref{#1}}}
\newcommand{\mainsec}[1]{\hyperref[#1]{Sec.~\ref{#1}}}
\newcommand{\tableref}[1]{\hyperref[#1]{Tab.~\ref{#1}}}

\newcommand{\supltableref}[1]{\hyperref[#1]{Suppl.~Tab.~\ref{#1}}}
\newcommand{\equationref}[1]{\hyperref[#1]{Eq.~\ref{#1}}}
\newcommand{\figureref}[1]{\hyperref[#1]{Fig.~\ref{#1}}}
\newcommand{\reffigures}[1]{\hyperref[#1]{Figures~\ref{#1}}}

\def\rvz{{\mathbf{z}}}

\DeclareMathAlphabet{\mathsfit}{\encodingdefault}{\sfdefault}{m}{sl}
\SetMathAlphabet{\mathsfit}{bold}{\encodingdefault}{\sfdefault}{bx}{n}

\def\gD{{\mathcal{D}}}

\def\gP{{\mathcal{P}}}

\definecolor{RegRow}{HTML}{FFF4C1}
\definecolor{RegBase}{HTML}{FFE08A}

\newcommand{\sectionref}[1]{\hyperref[#1]{Section~\ref{#1}}}

\definecolor{jointcolbg}{HTML}{F3F1F8}
\definecolor{standardcolbg}{HTML}{FAF5EC}

\newcommand{\Th}[1]{\textsc{#1}}

\newcommand{\mc}[2]{\multicolumn{#1}{c}{#2}}

\newcommand{\citeme}[1]{\red{[XX]}}
\newcommand{\refme}[1]{\red{(XX)}}

\newcommand{\decoder}{\Th{Decoder}\xspace}

\makeatletter
\newcommand*\bdot{\mathpalette\bdot@{.7}}
\newcommand*\bdot@[2]{\mathbin{\vcenter{\hbox{\scalebox{#2}{$\m@th#1\bullet$}}}}}
\makeatother

\makeatletter
\DeclareRobustCommand\onedot{\futurelet\@let@token\@onedot}
\def\@onedot{\ifx\@let@token.\else.\null\fi\xspace}

\makeatother
\usepackage{url}
\usepackage{tikz}
\usetikzlibrary{arrows.meta,positioning,calc,fit,backgrounds,shapes.misc,shapes.geometric}
\usepackage{xspace}
\usepackage{amsmath,amsfonts}
\usepackage{pifont}
\newcommand{\cmark}{\ding{51}}

\usepackage{multirow}
\usepackage{wrapfig,pgfplots} \pgfplotsset{compat=1.18}
\usepackage{xcolor}
\usepackage{colortbl}
\definecolor{spteal}{HTML}{1B8A7A}

\usepackage{multibib}
\newcites{supp}{Supplementary References}

\definecolor{cvprblue}{rgb}{0.21,0.49,0.74}
\usepackage[pagebackref,breaklinks,colorlinks,allcolors=cvprblue]{hyperref}

\def\confName{CVPR}
\def\confYear{2027}

\title{It's Not Too Late to Boost Your Latents:\\Semantic-Aware Decoding for Latent Generative Models}

\author{Giorgos~Petsangourakis$^{1,2}$ \hspace{0.5em}
Giorgos~Sfikas$^{2}$ \hspace{0.5em} Ioannis~Kakogeorgiou$^{1}$ \vspace{0.5em} 
\\
$^1$IIT, National Centre for Scientific Research “Demokritos” \hspace{1.0em} $^2$University of West Attica
}

\begin{document}

\twocolumn[{%
\renewcommand\twocolumn[1][]{#1}%
\maketitle

\addtocontents{toc}{\protect\setcounter{tocdepth}{-1}}

\begin{center}
    \captionsetup{type=figure}
    \resizebox{\textwidth}{!}{%
        \begingroup%
\definecolor{ink}{HTML}{263444}%
\definecolor{muted}{HTML}{8A97A6}%
\definecolor{genfill}{HTML}{EEF2F6}%
\definecolor{genedge}{HTML}{C3CDD8}%
\definecolor{latent}{HTML}{3F6FB5}%
\definecolor{latentfill}{HTML}{E6EFFC}%
\definecolor{semantic}{HTML}{E0823D}%
\definecolor{semanticfill}{HTML}{FDEFE3}%
\definecolor{decoderA}{HTML}{1B8A7A}%
\definecolor{decoderB}{HTML}{176A61}%
\definecolor{oursfill}{HTML}{F2FAF7}%
\definecolor{discard}{HTML}{C0674F}%

\newcommand{\sadfigzr}{\mathbf{z}_r}%
\newcommand{\sadfigzs}{\mathbf{z}_s}%
\newcommand{\sadfigzsp}{\hat{\mathbf{z}}_s}%
\newcommand{\sadfigxh}{\hat{\mathbf{x}}}%
\newcommand{\sadfigDsem}{\mathcal{D}_{\mathrm{sem}}}%
\newcommand{\sadfigProj}{\mathcal{P}}%

\begin{tikzpicture}[
  x=1cm,y=1cm,
  font=\sffamily\fontsize{9}{11}\selectfont,
  text=ink,
  >={Stealth[length=5pt,width=4pt,round]},
  gen/.style={
    draw=genedge,line width=.5pt,rounded corners=9pt,
    top color=white,bottom color=genfill,
    align=center,inner sep=6pt,
    preaction={fill=black,opacity=.05,transform canvas={yshift=-2.2pt}},
    preaction={fill=black,opacity=.05,transform canvas={yshift=-1pt}}
  },
  pill/.style={
    rounded rectangle,line width=.8pt,
    minimum width=1.1cm,minimum height=.56cm,inner xsep=6pt,
    font=\fontsize{12}{14}\selectfont
  },
  zr/.style={pill,draw=latent,fill=latentfill,text=latent},
  zs/.style={pill,draw=semantic,fill=semanticfill,text=semantic!80!black},
  plaindec/.style={
    draw=genedge,line width=.5pt,rounded corners=12pt,
    top color=white,bottom color=genfill,
    text width=2.1cm,minimum height=1.55cm,align=center,inner sep=6pt
  },
  semdec/.style={
    rounded corners=12pt,draw=none,
    left color=decoderA,right color=decoderB,text=white,
    text width=2.1cm,minimum height=1.9cm,align=center,inner sep=6pt,
    preaction={fill=decoderB,opacity=.10,transform canvas={yshift=-4pt}},
    preaction={fill=decoderB,opacity=.12,transform canvas={yshift=-2pt}}
  },
  proj/.style={
    trapezium,trapezium angle=72,shape border rotate=90,trapezium stretches body,
    draw=semantic,line width=.8pt,fill=white,rounded corners=2pt,
    minimum height=.62cm,minimum width=.62cm,inner sep=1pt,
    font=\fontsize{11}{12}\selectfont
  },
  flow/.style={->,line width=1.05pt,line cap=round},
  lflow/.style={flow,draw=latent},
  sflow/.style={flow,draw=semantic},
  oflow/.style={flow,draw=ink!80},
  note/.style={font=\sffamily\itshape\fontsize{8}{9}\selectfont},
  head/.style={font=\sffamily\bfseries\fontsize{10.5}{12}\selectfont},
  tag/.style={rounded rectangle,inner xsep=7pt,inner ysep=2.5pt,
    font=\sffamily\bfseries\fontsize{8}{9}\selectfont},
  formula/.style={font=\fontsize{10}{12}\selectfont,text=ink!85}
]
\path[use as bounding box] (-.35,-1.25) rectangle (16.75,6.75);

\newcommand{\sadfigout}[2]{

  \begin{scope}[shift={(#1)}]
    \node[rounded corners=4pt,minimum size=.78cm,inner sep=0,
          top color=latentfill,bottom color=semanticfill,
          draw=genedge,line width=.5pt] (#2) at (0,0) {};
    \clip[rounded corners=4pt] (-.39,-.39) rectangle (.39,.39);
    \fill[latent!55] (-.39,-.39) -- (-.12,.02) -- (.05,-.16) -- (.18,-.02) -- (.39,-.30) -- (.39,-.39) -- cycle;
    \fill[semantic!80] (.17,.20) circle (.075);
  \end{scope}
  \node[font=\fontsize{12}{14}\selectfont,below=1pt] at (#2.south) {$\sadfigxh$};
}

\begin{scope}[yshift=-0.3cm]

\node[head] at (3.9,6.78) {Joint generative models};
\node[head] at (12.45,6.78) {Standard latent generative models};

\node[tag,fill=genfill,text=muted,anchor=west] at (-.1,6.02) {Conventional decoding};

\begin{scope}[opacity=.72,transparency group]

\node[gen,minimum width=1.7cm,minimum height=1.55cm] (g1) at (.85,4.5) {Joint\\generator};
\node[zr] (r1) at (2.95,5.0) {$\sadfigzr$};
\node[zs,draw=semantic!45,fill=semanticfill!60,text=semantic!55,dash pattern=on 2pt off 1.5pt] (s1) at (2.95,4.0) {$\sadfigzs$};
\node[note,text=latent] at (2.95,5.52) {latents};
\node[plaindec] (d1) at (5.5,4.5) {Latents-only\\decoder\\[3pt]{\fontsize{13}{15}\selectfont$\mathcal D$}};
\draw[lflow] (g1.east) ++(0,.12) to[out=0,in=180] (r1.west);
\draw[sflow,draw=semantic!45] (g1.east) ++(0,-.12) to[out=0,in=180] (s1.west);
\draw[lflow] (r1.east) to[out=0,in=180] (d1.west);
\draw[draw=discard,line width=.9pt,dash pattern=on 1.5pt off 2pt,line cap=round]
  (s1.east) to[out=0,in=180] (4.05,3.8);
\draw[draw=discard,line width=1.2pt,line cap=round] (4.1,3.8) ++(-.09,-.09) -- ++(.18,.18) ++(0,-.18) -- ++(-.18,.18);
\node[note,text=discard,anchor=north] at (4.1,3.8) {discarded};
\sadfigout{7.7,4.5}{o1}
\draw[oflow] (d1.east) -- (o1.west);

\begin{scope}[xshift=8.55cm]
\node[gen,minimum width=1.6cm,minimum height=.95cm] (g2) at (.85,4.5) {Latent\\generator};
\node[zr] (r2) at (2.75,4.5) {$\sadfigzr$};
\node[note,text=latent] at (2.75,5.02) {latents};
\node[plaindec] (d2) at (5.5,4.5) {Latents-only\\decoder\\[3pt]{\fontsize{13}{15}\selectfont$\mathcal D$}};
\draw[lflow] (g2.east) -- (r2.west);
\draw[lflow] (r2.east) -- (d2.west);
\sadfigout{7.7,4.5}{o2}
\draw[oflow] (d2.east) -- (o2.west);
\end{scope}
\end{scope}
\end{scope}

\fill[oursfill,rounded corners=14pt,
      preaction={fill=black,opacity=.035,transform canvas={yshift=-3pt}},
      preaction={fill=black,opacity=.035,transform canvas={yshift=-1.5pt}}]
  (-.3,-.55) rectangle (16.7,2.95);
\node[tag,left color=decoderA,right color=decoderB,text=white,anchor=west] at (-.1,2.62)
  {Semantic-aware decoding (ours)};

\node[gen,minimum width=1.7cm,minimum height=1.55cm] (g3) at (.85,1.0) {Joint\\generator};
\node[zr] (r3) at (2.95,1.55) {$\sadfigzr$};
\node[zs] (s3) at (2.95,.45) {$\sadfigzs$};
\node[note,text=latent] at (2.95,2.07) {latents};
\node[note,text=semantic!85!black] at (2.95,0.97) {semantics};
\node[semdec] (d3) at (5.5,1.0)
  {\mbox{Semantic-aware}\\decoder\\[4pt]{\fontsize{13}{15}\selectfont$\sadfigDsem$}};
\draw[lflow] (g3.east) ++(0,.12) to[out=0,in=180] (r3.west);
\draw[sflow] (g3.east) ++(0,-.12) to[out=0,in=180] (s3.west);
\draw[lflow] (r3.east) to[out=0,in=180] ($(d3.west)+(0,.3)$);
\draw[sflow] (s3.east) to[out=0,in=180] ($(d3.west)+(0,-.3)$);
\sadfigout{7.7,1.0}{o3}
\draw[oflow] (d3.east) -- (o3.west);
\node[formula] at (2.9,-.2){$\sadfigxh=\sadfigDsem(\sadfigzr,\sadfigzs)$};

\begin{scope}[xshift=8.55cm]
\node[gen,minimum width=1.6cm,minimum height=.95cm] (g4) at (.85,1.75) {Latent\\generator};
\node[zr] (r4) at (2.45,1.75) {$\sadfigzr$};
\node[note,text=latent] at (2.45,2.27) {latents};
\node[proj] (p4) at (2.45,.4) {$\sadfigProj$};
\node[zs] (s4) at (3.4,.4) {$\sadfigzsp$};
\node[note,text=semantic!85!black] at (3.4,.92) {semantics};
\node[note,text=muted,anchor=east] at (p4.west) {projector};
\node[semdec] (d4) at (5.5,1.0)
  {\mbox{Semantic-aware}\\decoder\\[4pt]{\fontsize{13}{15}\selectfont$\sadfigDsem$}};
\draw[lflow] (g4.east) -- (r4.west);
\draw[lflow] (r4.east) to[out=0,in=180] ($(d4.west)+(0,.3)$);
\draw[lflow] (r4.south) to[out=-90,in=90] (p4.north);
\draw[sflow] (p4.east) -- (s4.west);
\draw[sflow] (s4.east) to[out=0,in=180] ($(d4.west)+(0,-.3)$);
\sadfigout{7.7,1.0}{o4}
\draw[oflow] (d4.east) -- (o4.west);
\node[formula] at (2.9,-.24)
  {$\sadfigxh=\sadfigDsem\bigl(\sadfigzr,\sadfigProj(\sadfigzr)\bigr)$};
\end{scope}

\draw[genedge!70,line width=.5pt,dash pattern=on 1pt off 2.5pt,line cap=round]
  (8.27,-.35) -- (8.27,6.25);
\end{tikzpicture}%
\endgroup
    }
    \vspace{-1.2cm}
    
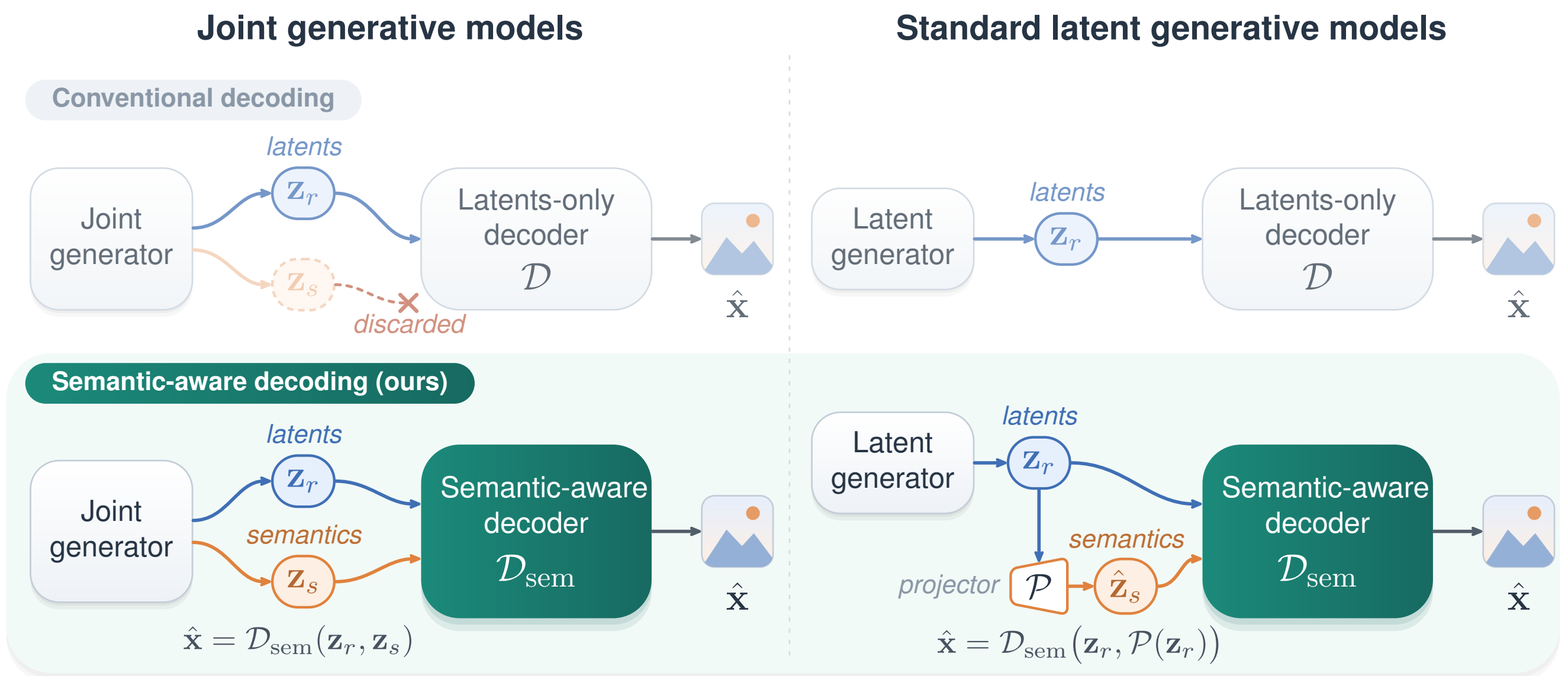
\captionof{figure}{\textbf{Semantic-aware decoding after sampling.}
    \textbf{Left:} joint generative models sample both latents $\mathbf{z}_r$ and semantics $\mathbf{z}_s$. \textbf{Right:} standard latent generative models sample only $\mathbf{z}_r$.
    \textbf{Top:} conventional decoding maps $\mathbf{z}_r$ to pixels with a
    latents-only decoder $\mathcal{D}$, discarding generated semantics when available.
    \textbf{Bottom:} our decoder $\mathcal{D}_{\mathrm{sem}}$ conditions on both
    latents and semantics, taken from the joint generator or, for standard models,
    predicted from latents by our projector as
    $\hat{\mathbf{z}}_s=\mathcal{P}(\mathbf{z}_r)$.
    The generator and \sdvae latent space remain unchanged.}
    \label{fig:teaser}
\end{center}
}]

\begin{abstract}

\vspace{-10pt}

Latent diffusion models increasingly exploit semantic representations from vision foundation models to improve image generation.
Yet models that jointly generate these representations alongside VAE latents discard them at decoding time, reconstructing images from the latents alone.
We introduce semantic-aware decoding, which conditions the decoder on both representations while keeping the pretrained generator and latent space fixed.
We further show that useful semantics need not be jointly generated: semantics extracted from a reconstruction and fed back to the semantic-aware decoder already improve generation, suggesting that conventional decoders do not fully exploit the information available in the latents.
Building on this observation, we train a semantic projector that predicts semantics directly from the latents, extending semantic-aware decoding to models that generate only VAE latents without an additional reconstruction at inference.
Experiments on ImageNet demonstrate improved generation across \texttt{SiT}, \texttt{REPA}, \texttt{REG}, \texttt{ReDi}, \texttt{CoReDi}, and \texttt{REGLUE}, spanning different model scales and both conditional and unconditional settings.
Semantic conditioning reduces generation gFID by up to 16\% over a strong latents-only transformer baseline.
Our findings show that semantic representations can improve image synthesis even after sampling is complete. The code is available at \href{https://github.com/giorgospets/semantic-aware-decoder}{https://github.com/giorgospets/semantic-aware-decoder}.
\end{abstract}    
\vspace{-6pt}
\section{Introduction}
\label{sec:intro}

Latent Diffusion Models (LDMs)~\citep{rombach2022high} are widely adopted for image generation. Rather than denoising pixels, they run the diffusion process in the compact latent space of a pretrained VAE~\citep{kingma2014}. The VAE \emph{encoder} maps images to latents, a diffusion model (the \emph{generator}) learns to model the distribution of these latents, and the VAE \emph{decoder} maps sampled latents back to pixels. The quality of the generated image therefore depends on the latent representation, how well the generator models it, and how effectively the decoder translates sampled latents back to pixel space.

Among these components, much of the recent progress has focused on the generator. A particularly promising direction is to explicitly encourage the learning of semantic structure. This is motivated by the observation that, under a denoising loss alone, the generator receives no explicit semantic target and learns semantic structure slowly~\citep{yu2024representation}. Several works address this by bringing features from pretrained Vision Foundation Models (VFMs) into training, which speeds up convergence and improves sample quality~\citep{yu2024representation,wu2025representation,kouzelis2025boosting,petsangourakis2026reglue}. 
\texttt{REPA}~\citep{yu2024representation} introduces semantic supervision by aligning intermediate generator representations with VFM features, while more recent approaches such as \reg~\citep{wu2025representation}, \redi~\citep{kouzelis2025boosting}, \coredi~\citep{kouzelis2026coevolving}, and \reglue~\citep{petsangourakis2026reglue} go further by jointly generating semantic representations alongside the VAE latents. These frameworks differ in the semantics they generate, ranging from global features to compressed patch-level representations.

On the encoder side, a parallel line of work incorporates semantic structure directly into the latent representation to make it easier for the generator to model.
\vavae~\citep{yao2025reconstruction} aligns the latents with VFM features during tokenizer training, \repae~\citep{leng2025repae} tunes the VAE end-to-end together with the diffusion model, and \rae~\citep{zheng2026diffusion} replaces the VAE encoder with a frozen VFM. 
Since these approaches define a new latent space, the generator must be trained from scratch on the corresponding latent space. Decoder-side adaptations, in contrast, can preserve the latent space and remain compatible with existing generators, as in \texttt{\sdvae-FT-EMA}~\citep{sdvaeftema}.

Despite the increasing use of semantic representations in both generator training and latent-space design, their role at the decoding stage remains largely unexplored. This is particularly striking for joint image-feature models, which generate semantic representations alongside VAE latents but discard them at decoding, leaving the decoder to reconstruct the image from VAE latents alone (see~\figureref{fig:teaser}, top left).

This raises a question: \emph{can semantics remain useful after sampling is complete?} VAE latents are optimized for reconstruction, whereas VFM representations capture higher-level semantic structure. Conditioning the decoder on both can therefore provide complementary information. For joint models in particular, these semantics are already available, requiring no changes to the latent space or generator.

In this work, we show that semantics remain useful even after sampling is complete, and that exploiting them at decoding time can improve both reconstruction and generation quality.  We introduce \textit{semantic-aware decoding} (\figureref{fig:teaser}), where the decoder reconstructs images from both VAE latents and VFM semantics while keeping the latent space and generator fixed. For joint models such as \reg, \redi, \coredi, and \reglue, these semantics are already produced during sampling and can therefore be exploited at no additional cost. We instantiate this idea with a transformer-based decoder operating on the same \sdvae latent space and study the contribution of patch-level (local) and image-level (global) semantics. Importantly, its latents-only counterpart already provides a stronger drop-in replacement for the standard \sdvae decoder. Our semantic conditioning yields substantial gains in generation quality over this stronger baseline.

More surprisingly, the semantics do not need to be explicitly generated. We show that useful semantic representations can be approximated from the VAE latents themselves. Specifically, we decode the latents, extract VFM features from the reconstructed image, and feed these representations back to the semantic-aware decoder. We refer to these recovered representations as \emph{proxy semantics}. Despite being derived only from the VAE latents, proxy semantics improve generation quality over the latents-only baseline, even when a joint model's generated semantics are discarded. This suggests that VAE latents contain information that conventional decoding does not fully exploit. Crucially, this extends the benefits of semantic-aware decoding to models that do not generate semantics, such as \texttt{SiT}~\citep{ma2024sit} and \texttt{REPA}. Finally, to avoid the additional decoding and VFM passes required by proxy semantics, we train a \emph{semantic projector} that predicts semantic representations directly from the VAE latents (\figureref{fig:teaser}, bottom right).

Taken together, our results show that semantic information can improve generation even after sampling is complete, whether it is directly produced by a joint generator or recovered from the VAE latents themselves. This enables semantic-aware decoding while preserving both the latent space and the pretrained generator. In summary, our contributions are:

\begin{itemize}
\item We introduce \emph{semantic-aware decoding}, which reuses semantic representations generated alongside VAE latents by joint modeling frameworks to improve generation quality. These gains come at no additional sampling cost, while both the latent space and the pretrained generator remain unchanged.

\item We show that useful semantic representations can be approximated from VAE latents alone by decoding the latents and re-encoding the resulting image with a VFM. These \emph{proxy semantics} also improve generation over the latents-only baseline, suggesting that conventional decoding does not fully exploit the information in these latents. We further introduce a \emph{semantic projector} that predicts semantic representations directly from the latents, avoiding the additional passes required by proxy semantics. This extends semantic-aware decoding to standard generators such as \texttt{SiT} and \texttt{REPA}.

\item We demonstrate consistent improvements in generation quality across several recent latent generative models, including extensively trained XL-scale generators, and in both conditional and unconditional settings. We further analyze the contributions of local and global semantics, compare jointly generated semantics with those approximated from VAE latents, and study compressed and uncompressed representations.

\end{itemize}

\section{Related work}
\label{sec:related}

\paragraph{Representations in generative modeling.} Earlier two-stage approaches generate semantic representations to condition image synthesis~\citep{ramesh2022hierarchical,li2024return}. More recently, \texttt{REPA}~\citep{yu2024representation} introduced an auxiliary alignment objective between projected generator features and pretrained VFM features to improve diffusion training. Subsequent work extends this alignment to attention and restricts it to early training~\citep{wang2025haste}, while further analysis highlights the importance of target spatial structure~\citep{singh2026what}. \texttt{DDT}~\citep{Wang2026CVPR} uses a VFM-aligned condition encoder to guide a separate velocity decoder within the denoising network. A complementary direction uses pretrained representations as the generative latent space. \rae-style methods pair frozen representation encoders with trainable image decoders~\citep{zheng2026diffusion,tong2026scaling}, while related approaches add a lightweight residual encoder to capture fine detail~\citep{shi2025SVG}. Most closely related to our work, joint modeling frameworks generate VAE latents alongside VFM semantics~\citep{wu2025representation,kouzelis2025boosting,kouzelis2026coevolving,petsangourakis2026reglue, wang2026jodiffusion,Pan2026CVPR, bao23aICML, ghoshdegeorge2026dinoforcing}. \reg~\citep{wu2025representation} models a global semantic token, whereas \redi~\citep{kouzelis2025boosting} and \coredi~\citep{kouzelis2026coevolving} use patch-level features compressed through PCA or a jointly learned linear projection, respectively. \reglue~\citep{petsangourakis2026reglue} combines global semantics with nonlinearly compressed patch features. Although these joint models exploit semantics during denoising, their image decoders use only the VAE latents, discarding the generated semantics at decoding time. Our work reuses these semantics to guide decoding, improving generation quality while keeping the latent space and pretrained generator unchanged. 

\paragraph{Autoencoder design.} Autoencoder design affects generation through both the latent space it defines and the mapping from latents to images. Beyond increasing compression~\citep{chen2025deep}, recent methods seek representations better suited to generative modeling. Semantic structure is encouraged through VFM supervision~\citep{yao2025reconstruction,xiong2025gigatok}, masked modeling~\citep{chen2025masked}, caption conditioning~\citep{zha2025language}, or adaptation of pretrained visual encoders~\citep{chen2026ICLR}. Other approaches promote equivariance~\citep{kouzelis2025eqvae,skorokhodov2025improving} and robustness to latent corruption~\citep{yang2026latent}, while \texttt{REPA-E} jointly tunes the autoencoder and generator~\citep{leng2025repae}. These approaches primarily shape the representation on which the generator is trained. Decoder design offers a complementary route, including diffusion-based reconstruction~\citep{zhao2025epsilon}. Several approaches preserve the existing latent space through \sdvae decoder fine-tuning~\citep{sdvaeftema}, consistency-distilled diffusion decoding~\citep{betker2023improving}, or single-step diffusion decoding with frozen pretrained encoders~\citep{vallaeys2026ssdd}. Asymmetric VQGAN increases decoder capacity and adds image conditioning for editing~\citep{zhu2023designing}. More recent work targets the mismatch between reconstruction and generation latents through decoder adaptation using reconstruction-based supervision~\citep{fang2026latent} or distributional objectives~\citep{fang2026efficient}. We complement these advances by examining semantic information supplied at decoding time. We keep the latent space and pretrained generator fixed and show that VFM conditioning improves generation beyond a latents-only baseline.

\paragraph{Mapping between representation spaces.} The Platonic representation hypothesis suggests that independently trained models can develop shared representational structure~\citep{huh2024platonic}. Model stitching connects pretrained networks through learned mappings~\citep{lenc2015understanding,bansal2021revisiting}, while relative representations~\citep{moschella2023relative} and latent-space translation~\citep{maiorca2023latent} use corresponding anchors to relate independently learned spaces. Subsequent approaches use spectral geometry to align representation manifolds~\citep{fumero2024latent}, construct a shared reference space for multiple models~\citep{achara2026multiway}, or translate text embeddings without paired data~\citep{jha2025harnessing}. Lightweight adaptor heads map features from a single distilled backbone into the representation spaces of several VFMs~\citep{ranzinger2024radio}, while universal sparse autoencoders reconstruct different models' activations through a shared concept space~\citep{derpanis2025universal}. We explore representation translation for semantic-aware decoding using a semantic projector that predicts VFM features directly from reconstruction-optimized VAE latents. These features are supplied alongside the original latents to the image decoder, improving generation even when the generator does not produce semantics.
\section{Method}
\label{sec:method}

\subsection{Preliminaries}
\label{sec:preliminaries}

\paragraph{Autoencoder and VFM.}
A VAE encoder $\mathcal{E}$ maps an image $\mathbf{x}\in\mathbb{R}^{3\times H\times W}$ to a latent $\mathbf{z}_r=\mathcal{E}(\mathbf{x})\in\mathbb{R}^{D_r\times H_r\times W_r}$, and a decoder $\mathcal{D}$ maps latents back to pixels $\hat{\mathbf{x}}=\mathcal{D}(\mathbf{z}_r)$. Since both are trained for reconstruction, we refer to $\mathbf{z}_r$ as \emph{reconstruction-optimized latents}. A VFM $\mathcal{F}$ (e.g., \texttt{DINOv2}~\citep{oquab2023dinov2}) maps an image to a global image-level \cls token  $\mathbf{f}_{\text{cls}}\in\mathbb{R}^{D_f}$ and local patch-level representations $\mathbf{f}_{\text{patch}}\in\mathbb{R}^{D_s\times H_s\times W_s}$ that capture high-level semantics.

\paragraph{Standard and joint generation.}
A standard LDM trains a generator, e.g., \sit~\citep{ma2024sit}, on $\mathbf{z}_r$ alone: it learns to denoise $\mathbf{z}_{r,t}=\alpha_t\mathbf{z}_r+\sigma_t\boldsymbol{\epsilon}$, $\boldsymbol{\epsilon}\sim\mathcal{N}(\mathbf{0},\mathbf{I})$, and sampling yields only $\mathbf{z}_r$. 
Joint modeling frameworks additionally model \emph{semantic latents} $\mathbf{z}_s$ derived from $\mathcal{F}(\mathbf{x})$, defining $\mathbf{z}=(\mathbf{z}_r,\mathbf{z}_s)$, and thus the generator learns to denoise $\mathbf{z}_t=\alpha_t\mathbf{z}+\sigma_t\boldsymbol{\epsilon}$, so that sampling yields both $\mathbf{z}_r$ and $\mathbf{z}_s$.
Our study builds on four joint modeling frameworks, which differ in the choice of $\mathbf{z}_s$: \reg uses the global token, $\mathbf{z}_s=\mathbf{f}_{\text{cls}}$; \redi uses PCA-compressed patch features, $\mathbf{z}_s=\mathrm{PCA}(\mathbf{f}_{\text{patch}})$; \coredi learns a linear projection $\mathcal{L}$ of the patch features jointly with the generator, $\mathbf{z}_s=\mathcal{L}(\mathbf{f}_{\text{patch}})$; and \reglue combines local and global semantics, $\mathbf{z}_s=(\mathcal{S}(\mathbf{f}_{\text{patch}}),\mathbf{f}_{\text{cls}})$, with $\mathcal{S}$ its learned semantic compressor. We summarize these choices as $\mathbf{z}_s=\mathcal{C}(\mathcal{F}(\mathbf{x}))$, where $\mathcal{C}$ denotes the framework-specific selection and compression of VFM features. Both $\mathcal{C}$ and $\mathcal{F}$ are kept frozen in our experiments.

\subsection{Semantic-Aware Decoder}
\label{sec:semdec}

In both standard and joint generative frameworks, decoding relies only on the latent $\mathbf{z}_r$ (see~\figureref{fig:teaser}, top row). This design choice is particularly striking for joint models, where the generated sample is $\mathbf{z}=(\mathbf{z}_r,\mathbf{z}_s)$ and thus already contains an additional semantic representation $\mathbf{z}_s$ that is discarded at decoding time. %

We instead propose a \emph{semantic-aware decoder} that exploits the full generated representation:
\begin{equation}
\hat{\mathbf{x}}=\mathcal{D}_{\text{sem}}(\mathbf{z}_r,\mathbf{z}_s).
\label{eq}
\end{equation}

With the VAE encoder $\mathcal{E}$ frozen, we train $\mathcal{D}_{\text{sem}}$ to reconstruct real images $\mathbf{x}$ from their reconstruction latents $\mathbf{z}_r=\mathcal{E}(\mathbf{x})$ and semantic representations $\mathbf{z}_s=\mathcal{C}(\mathcal{F}(\mathbf{x}))$. The decoder is optimized by minimizing

\begin{equation}
\mathcal{L}_{\text{dec}}=\|\hat{\mathbf{x}}-\mathbf{x}\|_1+\lambda_{\text{perc}}\,\mathcal{L}_{\text{perc}}(\hat{\mathbf{x}},\mathbf{x})+\lambda_{\text{adv}}\,\mathcal{L}_{\text{adv}}(\hat{\mathbf{x}}),
\label{eq:dec}
\end{equation}
where $\mathcal{L}_{\text{perc}}$ is the LPIPS perceptual loss and $\mathcal{L}_{\text{adv}}$ uses a discriminator following \rae~\citep{zheng2026diffusion}. At inference, $\mathbf{z}_s$ is produced by the joint model during sampling, at no additional sampling cost and without modifying the generator.

We explore both CNN-based and Transformer-based architectures for semantic-aware decoding. We adopt the Transformer architecture as our default, as it offers greater flexibility in incorporating global and local semantics and achieves better generation and reconstruction performance in our experiments (\mainsec{sec:analysis}).

\paragraph{Fusion with semantics.}
For the transformer decoder, we consider two ways of fusing patch-level semantics with the latents: token-wise concatenation, which adds semantic tokens to the input sequence, and channel-wise concatenation, which combines features at matching spatial positions. 
Since token-wise concatenation increases the sequence length, and hence the decoding cost, for only a marginal improvement in \gfid, we use channel-wise concatenation for patch-level semantics (\supplsec{supp:fusion_strategy}).
For the single global token, which has no spatial position, we instead follow \reg and \reglue and use token-wise concatenation, with negligible additional decoding overhead.

\subsection{Beyond Joint Models}
\label{sec:nonjoint}
 
\paragraph{Proxy semantics.}
Non-joint models such as \sit generate only $\mathbf{z}_r$. A straightforward way to approximate semantics is to decode the latents, encode the reconstruction with the VFM, and optionally compress the resulting features:
\begin{equation}
\tilde{\mathbf{z}}_s=\mathcal{C}\big(\mathcal{F}(\mathcal{D}(\mathbf{z}_r))\big).
\label{eq:proxy}
\end{equation}
For $\mathcal{C}$, we consider two choices: the \reglue semantic compressor and the identity mapping. The former yields proxy semantics in the same representation space as those generated by \reglue, enabling a controlled comparison between generated and proxy semantics. The latter retains the uncompressed $768$-channel \dinovtwob patch features and \cls token and is our default choice for non-joint models, as discussed in \mainsec{sec:proxy} (\tableref{tab:compression}). In each case, we use a semantic-aware decoder trained on the corresponding representation.

However, $\tilde{\mathbf{z}}_s$ is extracted from the imperfect reconstruction produced by the latents-only decoder and thus inherits its reconstruction errors. We refer to $\tilde{\mathbf{z}}_s$ as \emph{proxy semantics}, since it is derived from $\mathbf{z}_r$ rather than generated jointly with it. Computing these \emph{proxy semantics} also incurs additional inference cost, requiring an extra decoding pass and a VFM forward pass.

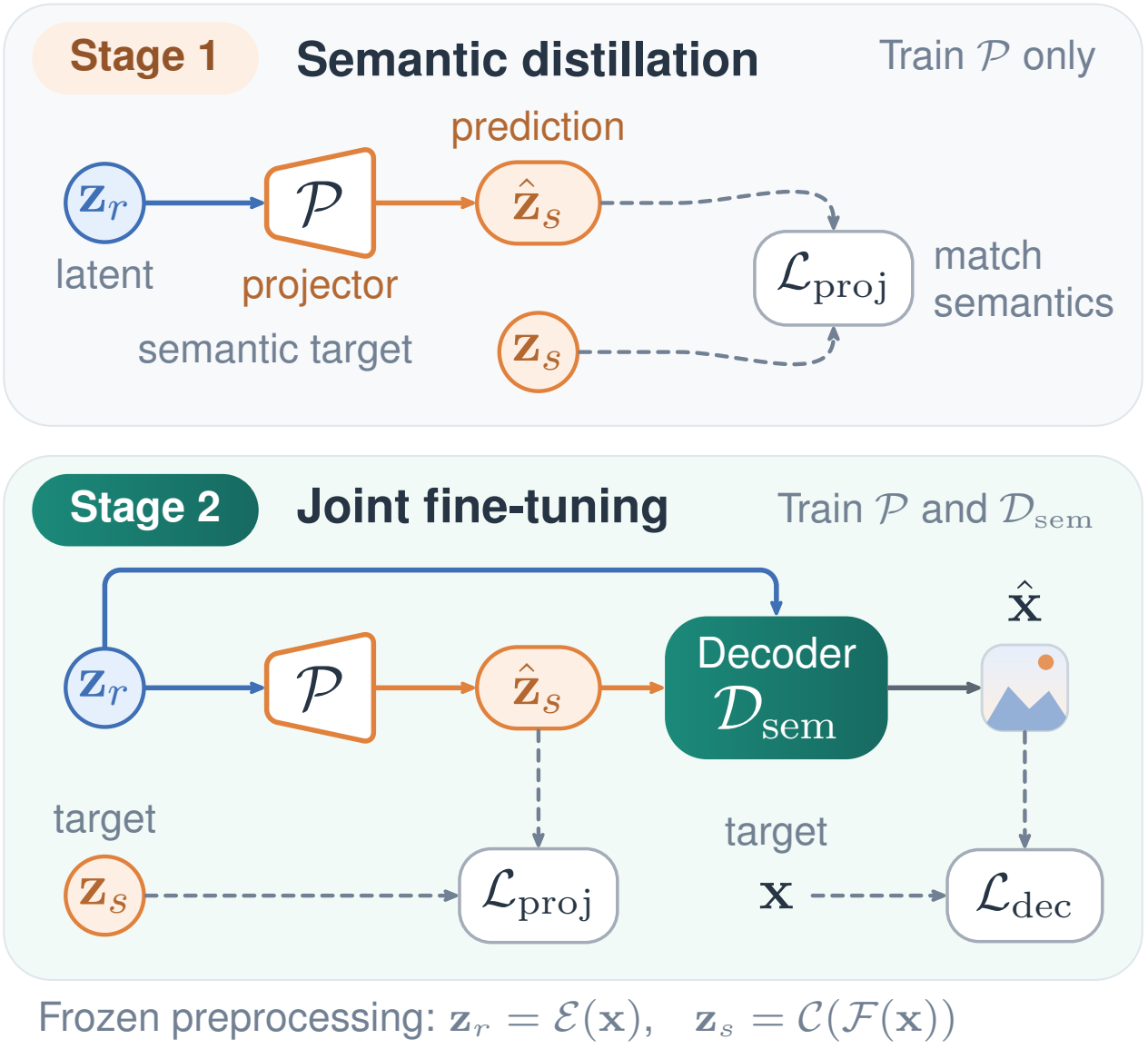
\begin{figure}
    \centering
    \resizebox{\linewidth}{!}{%
        \begingroup%
\definecolor{spink}{HTML}{263444}%
\definecolor{spmuted}{HTML}{718092}%
\definecolor{spedge}{HTML}{CCD5DE}%
\definecolor{spgray}{HTML}{F1F4F7}%
\definecolor{splatent}{HTML}{3F6FB5}%
\definecolor{splatentfill}{HTML}{E6EFFC}%
\definecolor{spsemantic}{HTML}{E0823D}%
\definecolor{spsemanticfill}{HTML}{FDEFE3}%
\definecolor{spteal}{HTML}{1B8A7A}%
\definecolor{sptealdark}{HTML}{176A61}%
\definecolor{sppanel}{HTML}{F2FAF7}%

\begin{tikzpicture}[
  x=1cm,y=1cm,
  font=\sffamily\fontsize{8}{9.6}\selectfont,text=spink,
  >={Stealth[length=4pt,width=3.2pt,round]},
  panel/.style={rounded corners=9pt,draw=spedge!60,line width=.35pt},
  pill/.style={rounded rectangle,minimum width=.78cm,minimum height=.52cm,
    inner xsep=5pt,inner ysep=3pt,line width=.75pt,
    font=\fontsize{11}{12}\selectfont},
  latent/.style={pill,draw=splatent,fill=splatentfill,text=splatent},
  semantic/.style={pill,draw=spsemantic,fill=spsemanticfill,text=spsemantic!80!black},
  projector/.style={
    trapezium,trapezium angle=72,shape border rotate=90,trapezium stretches body,
    rounded corners=2pt,draw=spsemantic,line width=.95pt,fill=white,
    minimum width=.66cm,minimum height=.71cm,inner sep=2pt,
    font=\fontsize{12}{13}\selectfont},
  decoder/.style={rounded corners=8pt,draw=none,
    left color=spteal,right color=sptealdark,text=white,
    text width=1.27cm,minimum height=.95cm,inner sep=3pt,align=center},
  loss/.style={rounded corners=6pt,draw=spmuted!65,line width=.6pt,
    fill=white,minimum width=1.03cm,minimum height=.61cm,inner sep=4pt,
    font=\fontsize{10}{12}\selectfont},
  flow/.style={->,line width=.95pt,line cap=round},
  lflow/.style={flow,draw=splatent},
  sflow/.style={flow,draw=spsemantic},
  oflow/.style={flow,draw=spink!75},
  supervise/.style={->,draw=spmuted,line width=.75pt,
    dash pattern=on 2pt off 1.7pt,line cap=round},
  title/.style={font=\sffamily\bfseries\fontsize{9}{11}\selectfont,anchor=west},
  spstage/.style={rounded rectangle,inner xsep=6pt,inner ysep=3pt,
    font=\sffamily\bfseries\fontsize{8}{9.6}\selectfont,anchor=west},
  small/.style={font=\sffamily\fontsize{7.6}{9}\selectfont,text=spmuted},
  sym/.style={font=\fontsize{11}{12}\selectfont}
]
\path[use as bounding box] (-.02,.48) rectangle (7.58,7.66);

\path[panel,fill=spgray!52] (0,4.82) rectangle (7.56,7.62);
\node[spstage,fill=spsemanticfill,text=spsemantic!65!black] (head1)
  at (.18,7.26) {Stage 1};
\node[title] at ($(head1.east)+(.12,0)$) {Semantic distillation};
\node[small,anchor=east] at (7.38,7.26) {Train $\mathcal P$ only};

\node[latent] (r1) at (.67,6.30) {$\mathbf z_r$};
\node[projector] (p1) at (2.10,6.30) {$\mathcal P$};
\node[semantic] (s1hat) at (3.55,6.30) {$\hat{\mathbf z}_s$};
\node[semantic] (s1) at (3.55,5.31) {$\mathbf z_s$};
\node[loss] (loss1) at (5.51,5.80) {$\mathcal L_{\mathrm{proj}}$};

\draw[lflow] (r1.east) -- (p1.west);
\draw[sflow] (p1.east) -- (s1hat.west);
\draw[supervise] (s1hat.east) -- (4.48,6.30)
  to[out=0,in=90] (loss1.north);
\draw[supervise] (s1.east) -- (4.48,5.31)
  to[out=0,in=-90] (loss1.south);

\node[small] at (.67,5.83) {latent};
\node[small,text=spsemantic!80!black] at (2.10,5.74) {projector};
\node[small,text=spsemantic!80!black] at (3.55,6.79) {prediction};
\node[small,anchor=east] at (2.84,5.31) {semantic target};
\node[small,anchor=west,align=left] at (6.05,5.80) {match\\semantics};

\path[panel,fill=sppanel] (0,1.14) rectangle (7.56,4.62);
\node[spstage,left color=spteal,right color=sptealdark,text=white] (head2)
  at (.18,4.26) {Stage 2};
\node[title] at ($(head2.east)+(.12,0)$) {Joint fine-tuning};
\node[small,anchor=east] at (7.38,4.26) {Train $\mathcal P$ and $\mathcal D_{\mathrm{sem}}$};

\begin{scope}[yshift=.30cm]
\node[latent] (r2) at (.67,2.78) {$\mathbf z_r$};
\node[projector] (p2) at (2.10,2.78) {$\mathcal P$};
\node[semantic] (s2hat) at (3.55,2.78) {$\hat{\mathbf z}_s$};
\node[decoder] (d2) at (5.13,2.78)
  {Decoder\\[2pt]{\fontsize{12}{13}\selectfont$\mathcal D_{\mathrm{sem}}$}};
\node[rounded corners=4pt,minimum size=.57cm,inner sep=0,
  top color=splatentfill,bottom color=spsemanticfill,draw=spedge,line width=.5pt]
  (out) at (6.78,2.78) {};
\begin{scope}
  \clip[rounded corners=4pt] (6.495,2.495) rectangle (7.065,3.065);
  \fill[splatent!55] (6.495,2.495) -- (6.67,2.79) -- (6.81,2.64) --
    (6.94,2.76) -- (7.065,2.57) -- (7.065,2.495) -- cycle;
  \fill[spsemantic!85] (6.92,2.95) circle (.052);
\end{scope}
\node[sym,anchor=south] at (6.78,3.10) {$\hat{\mathbf x}$};

\draw[lflow] (r2.east) -- (p2.west);
\draw[sflow] (p2.east) -- (s2hat.west);
\draw[sflow] (s2hat.east) -- (d2.west);
\draw[oflow] (d2.east) -- (out.west);

\coordinate (bypass2) at ($(d2.north)+(0,.30)$);
\draw[lflow,rounded corners=3pt]
  (r2.north) -- (r2.north |- bypass2)
  -- (bypass2) -- (d2.north);

\node[semantic] (s2) at (.67,1.40) {$\mathbf z_s$};
\node[small] at (.67,1.88) {target};
\node[loss] (loss2proj) at (3.55,1.40) {$\mathcal L_{\mathrm{proj}}$};
\draw[supervise] (s2.east) -- (loss2proj.west);
\draw[supervise] (s2hat.south) -- (loss2proj.north);

\node[loss] (loss2dec) at (6.78,1.40) {$\mathcal L_{\mathrm{dec}}$};
\node[sym] (xtarget) at (5.13,1.40) {$\mathbf x$};
\node[small] at (5.13,1.80) {target};
\draw[supervise] (out.south) -- (loss2dec.north);
\draw[supervise] (xtarget.east) -- (loss2dec.west);

\node[small,anchor=west] at (.10,.57)
  {Frozen preprocessing: $\mathbf z_r=\mathcal E(\mathbf x)$,\quad
   $\mathbf z_s=\mathcal C(\mathcal F(\mathbf x))$};
\end{scope}
\end{tikzpicture}%
\endgroup%
%
    }
    \vspace{-23pt}
    \caption{Two-stage training of the semantic projector.
    Dashed arrows indicate loss inputs.}
    \label{fig:projector-training}
\end{figure}

\paragraph{Semantic projector.}
To address these limitations, we train a semantic projector $\mathcal{P}$ to predict semantics directly from $\mathbf{z}_r$, bypassing the imperfect VAE reconstruction. We implement $\mathcal{P}$ using a ViT-B architecture and train it in two stages (\figureref{fig:projector-training}).

\textit{Stage-1: semantic distillation.} $\mathcal{P}$ learns to predict the semantics of an image from its latent:
\begin{equation}
\mathcal{L}_{\text{proj}}=\big\|\mathcal{P}\big(\mathcal{E}(\mathbf{x})\big)-\mathcal{C}\big(\mathcal{F}(\mathbf{x})\big)\big\|_2^2,
\label{eq:proj}
\end{equation}
summed over the patch-level and global components with $\mathcal{C}$ either of the two choices above. Unlike proxy semantics, the target comes from $\mathbf{x}$ itself rather than its reconstruction, and $\hat{\mathbf{z}}_s=\mathcal{P}(\mathbf{z}_r)$ needs no extra decoding.

\textit{Stage-2: joint fine-tuning.} Since Stage-1 does not involve the decoder, we jointly fine-tune $\mathcal{P}$ and $\mathcal{D}_{\text{sem}}$, initialized from Stage~1 and from the semantic-aware decoder of the target space, respectively. 
Training minimizes the combined objective $\mathcal{L}=\mathcal{L}_{\text{dec}}+\mathcal{L}_{\text{proj}}$, where the two loss terms are defined in \equationref{eq:dec} and \equationref{eq:proj}, respectively, and the reconstruction is given by $\hat{\mathbf{x}}=\mathcal{D}_{\text{sem}}(\mathbf{z}_r,\mathcal{P}(\mathbf{z}_r))$. Joint optimization allows the decoder and projector to adapt to each other while preserving semantic alignment (details in \supplsec{supp:projector}).
At inference, the same formulation decodes samples from non-joint models without modifying the generator. Beyond their practical value and performance gains, \emph{proxy semantics} and the \emph{semantic projector} provide an analytical tool for assessing whether and to what extent semantic-aware decoding depends on jointly generated semantics or can instead exploit semantic information derived from $\mathbf{z}_r$ alone. Unless otherwise stated, \emph{projected} refers to semantics by the jointly fine-tuned projector (Stage-2).

\section{Experiments}
\label{sec:experiments}

\subsection{Setup}
\label{sec:setup}

\paragnoskip{Implementation details.}

All decoders operate on latents $\rvz_r \in \mathbb{R}^{4 \times 32 \times 32}$ from the frozen pretrained \sdvae encoder. Our semantic-aware transformer decoder $\mathcal{D}_{\text{sem}}$ uses a \vitb backbone following the recipe of \rae, with one token per latent position. Each latent token predicts an $8 \times 8$ pixel patch, yielding a $256 \times 256$ image. Patch semantics are bilinearly upsampled to $32 \times 32$ and concatenated channel-wise with the latents, while the global \cls token is concatenated token-wise with the latent tokens. We train the decoder on \imagenet~\citep{deng2009imagenet} for $14$ epochs using $\ell_1$, LPIPS~\citep{zhang2018unreasonable}, and adversarial losses. To isolate the effect of semantic conditioning, we train a latents-only decoder $\mathcal{D}$ with the same backbone, objective, and training schedule.

For \reglue, we train three semantic-aware decoders conditioned on the \cls token, compressed patch features, or both. These representations are extracted from \dinovtwob with registers~\citep{darcet2024vision}, with patch features compressed by the \reglue semantic compressor to $16$ channels on a $16 \times 16$ spatial grid. For \reg, we use the \cls token of a register-free \dinovtwob, and for \redi, the first $8$ PCA components of \dinovtwob patch features. For \coredi, we use the $8$-channel output of its jointly learned linear projection of \dinovtwob (with registers) patch features; the projection is taken from the generator checkpoint and kept frozen. For uncompressed proxy and projected semantics, we train an additional semantic-aware decoder conditioned on the full $768$-channel patch features and \cls token of \dinovtwob with registers. The latents-only decoder is shared across all generators, while semantic-aware decoders can be reused wherever the semantic representation matches.

Proxy semantics are extracted from reconstructions of the latents-only decoder $\mathcal{D}$. For projected semantics, we train a \vitb projector $\mathcal{P}$ on \imagenet for $20$ epochs in Stage~1 using $\mathcal{L}_{\text{proj}}$. Both proxy and Stage-1 projected semantics use two semantic-aware decoders: the \reglue patch + \cls decoder for compressed semantics, or the decoder for uncompressed semantics. Stage~2 jointly fine-tunes each projector and its corresponding decoder for a further $2$ epochs using $\mathcal{L}_{\text{dec}}$ and $\mathcal{L}_{\text{proj}}$.

For the CNN experiments, we fine-tune a latents-only baseline and a semantic-aware variant from the original \sdvae~\citep{rombach2022high} for $5$ epochs, using the same objectives. The semantic-aware variant incorporates in patch semantics through a zero-initialized input convolution, whose output is summed with that of the pretrained one. We compare these with the commonly used \texttt{SD-VAE-FT-EMA} decoder.

For \texttt{B/2} generators, we use checkpoints trained for $400$K steps. At the \texttt{XL/2} scale, we use the most extensively trained checkpoints released by each framework: \sitxl at $7$M steps, \texttt{REPA-XL/2} at $4$M steps, \regxl at $4$M steps, \regluexl at $1$M steps, \redixl at $4$M steps, and \coredixl at $2$M steps. For unconditional generation, we train \regluexl for $400$K steps. For each frozen generator, all decoders use the same cached latent samples. We compare decoder performance within each generator, since generator architectures and training budgets differ across frameworks. 
Further architecture and training details are in \supplsec{supp:setup}.

\parag{Evaluation.}
We report Fr\'echet Inception Distance~\citep{heusel2017gans} for generation (\gfid) over $50$K generated samples and reconstruction (\rfid) on the $50$K-image \imagenet validation set. While our focus is on generation, \rfid provides a useful reference for further insights and analysis. For generation, semantics are jointly generated or recovered from the latents (\mainsec{sec:proxy}); for reconstruction, they are encoded from the input image. Additional metrics include sFID~\citep{nash2021generating}, Inception Score (IS)~\citep{salimans2016improved}, and Precision (Pre.) and Recall (Rec.)~\citep{kynkaanniemi2019improved} for generation, and PSNR, SSIM~\citep{wang2004image}, and LPIPS~\citep{zhang2018unreasonable} for reconstruction. The main tables report \gfid and \rfid; the full metric set (\supplsec{supp:metrics}), decoding costs (\supplsec{supp:additional_decoder_results}), and paired qualitative comparisons (\supplsec{sec:supp_vis}) are in the Supplementary. Generation metrics follow the standard ADM evaluation suite~\citep{dhariwal2021adm}. We generate samples using $250$-step Euler-Maruyama SDE sampling without classifier-free guidance.

\subsection{Analyzing semantic-aware decoding}
\label{sec:analysis}
\begin{table}[t]
\centering
\small
\caption{\textbf{Semantic-aware decoding on \regluebase and \regbase.}
All variants decode the same \texttt{SD-VAE} latents $\rvz_r$; the generator
and latent space remain unchanged.
\cmark\ marks semantics fed to decoder as extra input, generated jointly
with $\rvz_r$ for \gfid\ (at 400K iterations) and encoded from the input image for \rfid.
\textbf{(a)} is the \sdvaeftema decoder; $\dagger$: \textbf{(b)} and \textbf{(c)} are finetuned from original SD-VAE.
\textbf{(d)}-\textbf{(g)} share the same transformer architecture
and differ only in semantic input.
\textbf{(h)}-\textbf{(j)} repeat these configurations on \regbase,
whose semantics are global only.
\cls: global semantics; Patch-level: \reglue compressed semantics.}
\vspace{-6pt}
\label{tab:decoders_base}

\setlength{\tabcolsep}{3pt}
\vspace{-2pt}

\begin{tabular}{llcccc}
    \toprule
    &\multirow{2}{*}{Decoder}
    & \multicolumn{2}{c}{Semantic-aware}
    & \multirow{2}{*}{\gfid$\downarrow$}
    & \multirow{2}{*}{\rfid$\downarrow$}
    \\
    \cmidrule(lr){3-4}
    & & \cls & Patch-level & & \\
    \midrule
    \multicolumn{6}{@{}l}{\regluebase} \\ \midrule
    (a)           & \multirow{3}{*}{CNN} & --     & --     & 12.90 & 0.73 \\
    (b)$^\dagger$ & & --     & --     & 13.29 & 0.66 \\
    (c)$^\dagger$ & & \hl--     & \hl\cmark & \hl12.05 & \hl0.60 \\
    \cmidrule(l){2-6}
    (d) & \multirow{4}{*}{Transformer} & --     & --     & 12.08 & 0.41 \\
    (e) &                              & \hl\cmark & \hl--     & \hl10.91 & \hl0.36 \\
    (f) &                              & \hl--     & \hl\cmark & \hl10.42 & \hl0.35 \\
    (g) &                              & \hl\cmark & \hl\cmark & \hl\textbf{10.13} & \hl\textbf{0.34} \\
    \midrule
    \multicolumn{6}{@{}l}{\regbase} \\ \midrule
    (h) & CNN                          & --     & --     & 15.22 & 0.73 \\
    \cmidrule(l){2-6}
    (i) & \multirow{2}{*}{Transformer} & --     & --     & 14.15 & 0.41 \\
    (j) &                              & \hl\cmark & \hl-- & \hl\textbf{12.28} & \hl\textbf{0.34} \\
    \bottomrule
\end{tabular}
\vspace{-6pt}
\end{table}

We use \regluebase as our default testbed, since it jointly generates both global and local semantics (\tableref{tab:decoders_base}). We first incorporate semantics into the original convolutional \sdvae decoder~\citep{rombach2022high}, then show that a higher-capacity transformer decoder improves performance and benefits further from semantic conditioning. Using this more flexible architecture, we compare global, local, and combined semantics, before extending our evaluation to \reg (\tableref{tab:decoders_base}).

\parag{Semantic conditioning improves decoding.}
We first incorporate semantics into the original CNN-based \texttt{SD-VAE} decoder. Simply fine-tuning the convolutional decoder on \imagenet \textbf{(b)} degrades \gfid from $12.90$ \textbf{(a)} to $13.29$, despite improving \rfid from $0.73$ to $0.66$ (\tableref{tab:decoders_base}). In contrast, conditioning the same decoder on patch semantics \textbf{(c)} reduces \gfid to $12.05$, a $6.6\%$ improvement over \textbf{(a)}, while further improving  \rfid to $0.60$. Thus, semantics improve both generation and reconstruction quality.

\parag{A stronger transformer baseline benefits further from semantics.}
Replacing the convolutional decoder with the latents-only transformer decoder \textbf{(d)} improves \gfid from $12.90$ to $12.08$ and \rfid from $0.73$ to $0.41$, without using semantic representations. This improvement cannot be attributed to additional training, as shown by the results of the fine-tuned convolutional decoder \textbf{(b)}. We therefore adopt the latents-only transformer \textbf{(d)} as the stronger baseline. Conditioning this decoder on the semantics jointly generated by \regluebase further reduces \gfid from $12.08$ \textbf{(d)} to $10.13$ \textbf{(g)}, a $16.1\%$ reduction, while semantics encoded from the input image improve \rfid from $0.41$ to $0.34$. These gains require neither retraining nor modifying the generator and introduce no additional sampling cost, since the semantic representations are already produced jointly with the latents.

\parag{Local and global semantics work in synergy.}
Using only global semantics \textbf{(e)} reduces \gfid from $12.08$ to $10.91$ and \rfid from $0.41$ to $0.36$. Patch semantics \textbf{(f)} provide a larger improvement, reaching $10.42$ \gfid and $0.35$ \rfid. Combining global and patch semantics \textbf{(g)} performs best, achieving $10.13$ \gfid and $0.34$ \rfid. Although most of the gain is driven by spatially structured patch semantics, the additional improvement from the global token indicates a beneficial interaction between local and global information.

\parag{The benefit extends to \texttt{REG}.}
We observe the same effect with \regbase (\tableref{tab:decoders_base}), which jointly generates a global semantic token. Conditioning on this generated token boosts \gfid from $14.15$ \textbf{(i)} to $12.28$ \textbf{(j)}, a $13.2\%$ improvement. This shows that the benefit is not specific to \reglue or to spatial semantic representations.

\begin{table}[t]
\centering
\small

\caption{\textbf{Semantic-aware decoding on XL generators.}
    \gfid$\downarrow$ of SiT-XL/2 backbone generators.
    All decoders operate on the same \sdvae latents $\rvz_r$;
    \textit{CNN} is the \sdvaeftema decoder.
    \textit{Semantic-aware}: the transformer decoder is conditioned on the semantics jointly generated with $\rvz_r$:
    patch-level and \cls for \reglue, \cls for \reg, and $8$-channel patch-level semantics for \redi and \coredi. \sit and \repa generate no semantics; these are first predicted from $\rvz_r$ by the semantic projector (after Stage-2) and then passed to the semantic-aware decoder.
    The generator and latent space remain unchanged.}
\vspace{-6pt}
\label{tab:xl_scale}
\setlength{\tabcolsep}{3pt}

\begin{tabular}{@{}lccc}
    \toprule
    \multirow{2}{*}{Generator}
    & \multirow{2}{*}{CNN}
    & \multicolumn{2}{c}{Transformer} \\
    \cmidrule(l){3-4}
    & & Latents-only & Semantic-aware\\
    \midrule
    \mc{4}{(a) Joint generative models }\\

\arrayrulecolor{black!30}\cmidrule(lr){1-4}\arrayrulecolor{black}
    \coredi           & 3.30 & 3.06 & \hl\textbf{2.79} \\
    \redi             & 3.30 & 2.85 & \hl\textbf{2.65} \\
    \reglue           & 2.50 & 2.43 & \hl\textbf{2.15} \\
    \reg              & 1.80 & 1.63 & \hl\textbf{1.54} \\

    \midrule
    \mc{4}{(b) Standard latent generative models}\\
\arrayrulecolor{black!30}\cmidrule(lr){1-4}\arrayrulecolor{black}
    \sit & \phantom{$\dagger$}8.60$^\dagger$ & 7.58 & \hl\textbf{6.58} \\
    
    \repa & 5.90 & 4.95 & \hl\textbf{4.19} \\
    \bottomrule
    \\[-8pt]
    \multicolumn{4}{@{}c}{\small{$\dagger$: our reproduction; 8.30 reported in~\citep{ma2024sit}}} \\[-2pt]
    
\end{tabular}
\vspace{-6pt}
\end{table}
\begin{figure}[t]
  \centering
  \includegraphics[width=\linewidth]{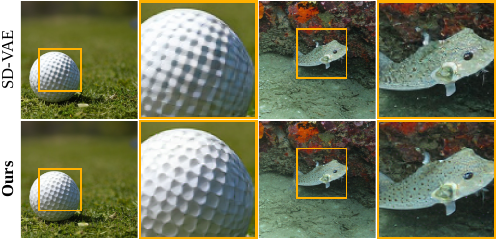}
  \captionsetup{skip=3pt}
  \caption{
  \textbf{Qualitative comparison.} \regluexl samples decoded from the same latents by the \sdvaeftema decoder (top) and by our semantic-aware decoder with \emph{jointly generated} semantics (bottom).}
  \label{fig:qualitative}
\vspace{-6pt}
\end{figure}
\parag{XL generators.}
\tableref{tab:xl_scale}\hspace{-3pt}~\textcolor{cvprblue}{(a)} extends the analysis at XL scale. Again, using the jointly generated semantics during decoding yields the best performance across all settings: on \regluexl, global and local semantics boost \gfid from $2.43$ to $2.15$ ($11.5$\% improvement), and on \regxl, the \cls token improves performance from $1.63$ to $1.54$ \gfid ($5.5$\%). The qualitative comparison in \figureref{fig:qualitative} shows that, for the same \regluexl latents, semantic-aware decoding produces clearer golf-ball dimples and fish markings than standard latents-only decoding. We broaden our evaluation to include two more recent approaches that jointly generate $8$-channel patch semantics. Semantic-aware decoding enhances \gfid\ from $2.85$ to $2.65$ ($7.0$\%) and from $3.06$ to $2.79$ ($8.8$\%) for \redixl and \coredixl, respectively.
Additional experiments with classifier-free guidance show that semantic-aware decoding improves or matches \gfid (see ~\supltableref{tab:cfg_results}).

\parag{Semantics also improve unconditional generation.} 
\tableref{tab:decoders_reglue_uncond} shows that semantic-aware decoding also improves unconditional generation on \regluebase and \regluexl, both trained for $400$K iterations. Conditioning the decoder on jointly generated semantics improves \gfid from $26.25$ to $23.68$ ($9.8$\%) on \regluebase and from $8.63$ to $7.52$ ($12.9$\%) on \regluexl, relative to the latents-only transformer. We use the same semantic-aware decoder as in the class-conditional experiments.

\begin{table}[t]
\centering
\caption{\textbf{Semantic-aware decoding for unconditional generation.}
\gfid$\downarrow$ on \reglue unconditional generators trained for $400$K iterations.
All decoders use the same \texttt{SD-VAE} latents; semantic-aware decoding also uses jointly generated local and global semantics.
\textit{CNN} denotes the \sdvaeftema decoder.}
\vspace{-6pt}
\label{tab:decoders_reglue_uncond}
\small
\setlength{\tabcolsep}{5pt}
\begin{tabular}{@{}lccc}
    \toprule
    \multirow{2}{*}{Generator}
    & \multirow{2}{*}{\textit{CNN}}
    & \multicolumn{2}{c}{Transformer} \\
    \cmidrule(l){3-4}
    & & Latents-only & Semantic-aware \\
    \midrule
    \regluebase  & 27.95 & 26.25 & \hl\textbf{23.68} \\
    \regluexl & \xspace\xspace9.47  & \xspace\xspace8.63  & \xspace\xspace\hl\textbf{7.52} \\
    \bottomrule
\end{tabular}
\vspace{-6pt}
\end{table}

\subsection{Decoding without generated semantics}
\label{sec:proxy}

{We next evaluate semantic-aware decoding with \emph{proxy} and \emph{projected} semantics derived from $\rvz_r$ alone (~\mainsec{sec:nonjoint}). Our evaluation covers joint generators as well as \sit and \repa, which do not generate semantics.

\parag{The effect of semantic compression.}
\tableref{tab:compression} compares compressed and uncompressed \dinovtwob patch features, with the \cls token included in both settings. Compression has mixed effects on proxy semantics, improving \gfid on \sitxl but slightly degrading it on \regluexl. For projected semantics after joint fine-tuning (Stage~2), uncompressed features improve \gfid from $6.76$ to $6.58$ on \sitxl and from $2.26$ to $2.20$ on \regluexl. This trend holds across the other generators (~\supltableref{tab:supp_gen_xl_proxy_proj}). We therefore adopt uncompressed patch features and the \cls token as our default representation for proxy and projected semantics.

\begin{table}[t]
\centering
\caption{\textbf{Compressed vs. uncompressed patch semantics for semantic-aware decoding.}
\gfid$\downarrow$ on XL/2 generators.
Compression uses the \reglue compressor on \dinovtwob patch features; both settings include the uncompressed \cls token.
Proxy and projected semantics are derived from $\rvz_r$ alone, with projected results reported after joint fine-tuning (Stage-2).
Latents-only refers to the transformer baseline; the generated \reglue reference uses compressed patch semantics by design.}
\vspace{-6pt}
\label{tab:compression}
\small
\setlength{\tabcolsep}{8pt}
\begin{tabular}{@{}lcc@{}}
    \toprule
    Representation & Proxy & Projected\\
    \midrule
    \multicolumn{3}{@{}l}{
        \sitxl\quad\quad\textit{(latents-only: $7.58$; generated: \redx)}
    } \\
    \cmidrule{1-3}
    Compressed   & 6.83 & 6.76 \\
    Uncompressed & 7.37 & \textbf{6.58} \\
    \midrule
    \multicolumn{3}{@{}l}{
        \regluexl\xspace\textit{(latents-only: $2.43$; generated: $2.15$)}
    } \\
    \cmidrule{1-3}
    Compressed   & 2.32 & 2.26 \\
    Uncompressed & 2.30 & \textbf{2.20} \\
    \bottomrule
\end{tabular}
\vspace{-6pt}
\end{table}

\parag{Proxy semantics improve generation.}
We show that proxy semantics improve \gfid over the latents-only decoder by $2.8$-$6.2$\% across all six generators (\tableref{tab:sem_sources_reglue_reg}). This includes \sit and \repa, which do not generate semantics, where \gfid improves from $7.58$ to $7.37$ and from $4.95$ to $4.70$, respectively. Since proxy semantics are derived from $\rvz_r$ alone, these gains show that making semantic information explicit through the VFM benefits decoding without requiring joint generation. Despite outperforming latents-only decoding, proxy semantics remain behind the generated references on all four joint generators. We argue that this gap partly reflects reconstruction errors inherited by proxy semantics.

\begin{table}[t]
\centering
\caption{\textbf{Decoding with proxy and projected semantics.}
\gfid$\downarrow$ on XL/2 generators.
Proxy semantics are extracted from latents-only reconstructions; projected semantics are predicted directly from $\rvz_r$.
Both use uncompressed $768$-channel \dinovtwob patch features and the \cls token.
Stage~1 uses a frozen decoder; Stage~2 jointly fine-tunes the projector and decoder.
Generated references use each framework's own representation and matching decoder.
Latents-only denotes the transformer baseline.
Bold marks the best result without generated semantics.}
\vspace{-6pt}
\label{tab:sem_sources_reglue_reg}
\small
\setlength{\tabcolsep}{2.pt}
\begin{tabular}{@{}lccccc}
    \toprule
    \multirow{2}{*}{Generator}
    & \multirow{2}{*}{\textcolor{gray}{Generated}}
    & \multirow{2}{*}{Latents-only}
    & \multirow{2}{*}{Proxy}
    & \multicolumn{2}{c}{Projected} \\
    \cmidrule(lr){5-6}
    & & & & Stage~1 & Stage~2 \\
    \midrule
    \sit
        & \redx & 7.58 & 7.37 & 6.92 & \hl\textbf{6.58} \\
    \repa
        & \redx & 4.95 & 4.70 & 4.55 & \hl\textbf{4.19} \\
    \arrayrulecolor{black!30}
    \cmidrule(lr){1-6}
    \arrayrulecolor{black}
    \coredi
        & \textcolor{gray}{2.79}
        & 3.06 & 2.87 & 2.84 & \hl\textbf{2.64} \\
    \redi
        & \textcolor{gray}{2.65}
        & 2.85 & 2.71 & 2.68 & \hl\textbf{2.46} \\
    \reglue
        & \textcolor{gray}{2.15}
        & 2.43 & 2.30 & 2.38 & \hl\textbf{2.20} \\
    \reg
        & \textcolor{gray}{1.54}
        & 1.63 & 1.56 & 1.72 & \hl\textbf{1.54} \\
    \bottomrule
\end{tabular}
\vspace{-6pt}
\end{table}

\begin{table}[t]
\centering
\caption{\textbf{Decoder capacity, throughput, and generation quality.}
\gfid$\downarrow$ on XL/2 generators.
Base and Large are \vitb and \vitl latents-only decoders; semantic-aware decoding uses \vitb.
Projected uses uncompressed \dinovtwob patch features and the \cls token predicted by our semantic projector after joint fine-tuning with the decoder (Stage-2).
Generated uses each framework's jointly generated semantics.
Costs include the projector where used and exclude the generator.
Green shading marks semantic-aware decoding;
\redx~denotes unavailable generated semantics.
$^{*}$ Variations across joint frameworks are negligible ($<0.01$M parameters and $<0.1$ im/s).}
\vspace{-6pt}
\label{tab:decoder_params}
\small
\setlength{\tabcolsep}{4pt}
\begin{tabular}{@{}lcccc}
    \toprule
    \multirow{2}{*}{Decoder}
    & \multicolumn{2}{c}{Latents-only}
    & \multicolumn{2}{c}{Semantic-aware} \\
    \cmidrule(lr){2-3}\cmidrule(lr){4-5}
    & Base & Large
    & Projected & Generated \\
    \midrule
    \#Params (M)
        & 86.0 & 303.6 & \hl173.4 & \hl86.6$^{*}$ \\
    \begin{tabular}[c]{@{}l@{}}
        Throughput (im/s)
    \end{tabular}
        & 127.8 & 46.42 & \hl116.8 & \hl127.8$^{*}$ \\
    \midrule
    Generator & \multicolumn{4}{c}{\gfid$\downarrow$} \\
    \cmidrule(lr){2-5}
    \sit
        & 7.58 & 7.45 & \hl\textbf{6.58} & \redx \\
    \repa
        & 4.95 & 4.83 & \hl\textbf{4.19} & \redx \\
    \midrule
    \coredi
        & 3.06 & 2.98 & \hl\textbf{2.64} & \hl2.79 \\
    \redi
        & 2.85 & 2.81 & \hl\textbf{2.46} & \hl2.65 \\
    \reglue
        & 2.43 & 2.41 & \hl2.20 & \hl\textbf{2.15} \\
    \reg
        & 1.63 & 1.60 & \hl\textbf{1.54}
        & \hl\textbf{1.54} \\
    \bottomrule
\end{tabular}
\vspace{-6pt}
\end{table}

\parag{Predicting semantics directly from latents.}
The semantic projector is trained on semantic targets extracted from real images and predicts semantics directly from $\rvz_r$, avoiding the preliminary decoding and VFM passes required by proxy semantics. After Stage-1, projected semantics use the same frozen decoder as proxy semantics and improve \gfid on four of the six generators, but underperform on \reg and \reglue. Joint fine-tuning of the projector and decoder (Stage-2) improves \gfid over proxy semantics and Stage-1 across all six generators (\tableref{tab:sem_sources_reglue_reg}). On \reglue, \gfid improves from $2.38$ to $2.20$, outperforming proxy semantics ($2.30$). On \sit and \repa, Stage-2 achieves $6.58$ and $4.19$ \gfid, improving by $13.2$\% and $15.4$\% over latents-only decoding, respectively. Compared with official \sdvae results, our decoder improves \gfid on \sit ($8.30{\to}6.58$, $20.7$\%) and \repa ($5.90{\to}4.19$, $29.0$\%) (\tableref{tab:xl_scale}\hspace{-2pt}~\textcolor{cvprblue}{(b)}).

\parag{Decoder capacity and efficiency.}
Scaling the latents-only decoder from \vitb to \vitl results in more than three times as many parameters and lowers throughput from $127.8$ to $46.42$ img/s, while yielding only modest \gfid improvements (\tableref{tab:decoder_params}). With projected semantics, semantic-aware decoding outperforms \vitl across all six generators with $42.9$\% fewer parameters ($173.4$M versus $303.6$M) and $2.5\times$ higher throughput ($116.8$ versus $46.42$ img/s). Reusing generated semantics adds only $0.6$M parameters over the \vitb baseline without reducing throughput, while improving \gfid on all four joint generators. 
\supplsec{supp:additional_decoder_results} reports further decoder-efficiency results.

\setlength{\columnsep}{11pt}
\begin{wrapfigure}{r}{0.55\linewidth}
  \centering
  \vspace{-13pt}
  \begin{tikzpicture}
    \begin{axis}[
      ybar=5pt,
      scale only axis,
      width=0.88\linewidth,
      height=3.2cm,
      bar width=11pt,
      ymin=0, ymax=93,
      ytick={0,20,40,60,80},
      enlarge x limits=0.5,
      symbolic x coords={probe,miou},
      xtick=data,
      xticklabels={ImageNet\\probing acc., Cityscapes\\mIoU},
      xticklabel style={align=center, font=\scriptsize},
      yticklabel style={font=\scriptsize},
      major x tick style={draw=none},
      ymajorgrids,
      grid style={dashed, gray!30},
      nodes near coords,
      nodes near coords style={font=\tiny},
      legend style={
        at={(0.5,1.03)},
        anchor=south,
        legend columns=-1,
        draw=none,
        font=\scriptsize,
        column sep=4pt
      },
    ]
      \addplot[fill=gray!60, draw=none]
        coordinates {(probe,30.3) (miou,59.14)};
      \addplot[fill=blue!60, draw=none]
        coordinates {(probe,84.2) (miou,72.50)};
      \legend{\redi (PCA), Uncompressed}
    \end{axis}
  \end{tikzpicture}
  \vspace{-1.8em}
  \caption{Semantic quality of \redi's PCA-compressed
  ($8$-channel) and uncompressed ($768$-channel)
  \dinovtwob patch features.
  Attentive probing and segmentation results following
  \citep{petsangourakis2026reglue}.}
  \label{fig:redi_reglue_sem}
  \vspace{-1em}
\end{wrapfigure}
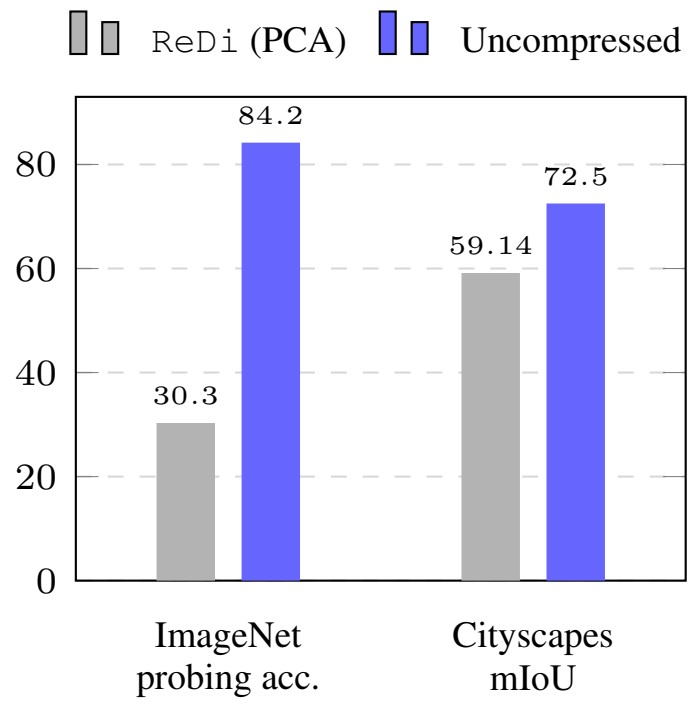

\parag{The role of semantic richness.}
On \redi and \coredi, projected semantics derived from the same $\rvz_r$ improve \gfid over generated semantics from $2.65$ to $2.46$ and from $2.79$ to $2.64$, respectively (\tableref{tab:sem_sources_reglue_reg}). Both frameworks generate compact $8$-channel patch semantics without a \cls token, obtained by PCA in \redi and a learned linear projection in \coredi. Our projector approximates the full $768$-channel \dinovtwob patch features and \cls token. Probing and segmentation results on encoded features show advantages in both class-level discrimination and dense spatial semantics (\figureref{fig:redi_reglue_sem}): uncompressed features outperform \redi's PCA features in both ImageNet probing accuracy ($84.2$\% vs.\ $30.3$\%) and Cityscapes mIoU ($72.50$ vs.\ $59.14$)~\citep{petsangourakis2026reglue}. We argue that this richer representation partly explains the gains, consistent with prior evidence that semantic richness can benefit generation~\citep{petsangourakis2026reglue,yu2024representation}. More discussion in \supplsec{supp:additional_results_richness}.
\section{Conclusion and Future Work}
\label{sec:conclusion}

We showed that semantic representations remain useful even after sampling is complete. By conditioning the decoder on VFM semantics, either jointly generated with the VAE latents or approximated from the latents themselves, we improve reconstruction and generation quality without modifying the pretrained generator or latent space. Our results show that semantics can expose information that conventional latents-only decoding does not fully exploit, establishing semantic-aware decoding as an effective post hoc mechanism for improving latent generators.

Future work can explore more efficient semantic projectors, alternative semantic representations, foundation models, and unified decoders that can flexibly exploit reconstruction-oriented latents, semantic representations, or both. It would also be interesting to study the trade-offs between semantic compression and its utility to the decoder.

\section{Acknowledgements}
\label{sec:acknowledgements}
We acknowledge the EuroHPC Joint Undertaking for awarding this project
access to the EuroHPC supercomputer LEONARDO, hosted by CINECA (Italy)
and the LEONARDO consortium through an EuroHPC Development Access call
(Projects ID EHPC-DEV-2026D02-285 \& EHPC-DEV-2026D03-137).
We also acknowledge the EuroHPC Joint Undertaking for awarding us access
to Karolina, hosted by IT4Innovations (Czech Republic), through an EuroHPC
Development Access call (project ID EHPC-DEV-2026D06-291).
{
    \small
     \bibliographystyle{ieeenat_fullname}
    \bibliography{main}
}

\addtocontents{toc}{\protect\setcounter{tocdepth}{2}}

\clearpage
\maketitlesupplementary

{
  \hypersetup{linkcolor=black}
  \tableofcontents
}

\setcounter{section}{0}
\renewcommand{\thesection}{\Alph{section}}
\renewcommand{\theHsection}{\Alph{section}}
\setcounter{table}{0}
\renewcommand{\thetable}{S\arabic{table}}
\renewcommand{\theHtable}{S\arabic{table}}
\setcounter{figure}{0}
\renewcommand{\thefigure}{S\arabic{figure}}
\renewcommand{\theHfigure}{S\arabic{figure}}
\setcounter{equation}{0}
\renewcommand{\theequation}{S\arabic{equation}}
\renewcommand{\theHequation}{S\arabic{equation}}

\providecommand{\vitl}{ViT-L\xspace}
\providecommand{\repaxl}{\texttt{REPA-XL/2}\xspace}
\providecommand{\tbd}{\todo{X.XX}\xspace}
\providecommand{\suppfigure}[3]{%
  \IfFileExists{#1}{\includegraphics[width=#2]{#1}}{%
    \fbox{\parbox[c][#3][c]{\dimexpr#2-2\fboxsep-2\fboxrule\relax}{\centering
      \footnotesize\color{red}Placeholder:\\\nolinkurl{#1}}}}}

\section{Additional Experimental Results}
\label{supp:results}

\subsection{Decoder Training Dynamics}
\label{supp:curves}

\figureref{fig:supp_curves} tracks \rfid and \gfid of the transformer decoders over their $14$ training epochs.
\rfid is computed on the \imagenet validation set with semantics encoded from the input image, and \gfid on \regluexl samples with jointly generated semantics (50k samples per evaluation).
Both metrics drop sharply at epoch $3$, when we introduce the adversarial loss.
From then on, all semantic-aware decoders outperform the latents-only decoder on both metrics. In \gfid, the latents-only decoder plateaus near \texttt{SD-VAE-FT-EMA}, while the semantic-aware decoders stay well below it.
In \gfid, local patch semantics help more than the global \texttt{[CLS]} token, and combining the two gives the best result throughout.
All results use the final checkpoint.

\pgfplotstableread{
epoch latents  cls   patch  clspatch
1     1.69     1.53  1.34   1.30
2     1.63     1.53  1.36   1.37
3     0.56     0.50  0.44   0.43
4     0.50     0.45  0.40   0.39
5     0.47     0.42  0.38   0.37
6     0.45     0.40  0.37   0.36
7     0.44     0.39  0.37   0.36
8     0.43     0.38  0.36   0.35
9     0.42     0.37  0.36   0.35
10    0.42     0.37  0.36   0.34
11    0.41     0.37  0.35   0.35
12    0.41     0.36  0.35   0.35
13    0.41     0.36  0.35   0.34
14    0.41     0.36  0.35   0.34
}\suppCurvesRfid

\pgfplotstableread{
epoch latents  cls   patch  clspatch
1     3.65     3.62  3.42   3.37
2     3.53     3.54  3.26   3.26
3     2.57     2.47  2.32   2.28
4     2.51     2.41  2.27   2.22
5     2.49     2.37  2.23   2.20
6     2.47     2.35  2.22   2.18
7     2.47     2.33  2.22   2.17
8     2.45     2.32  2.21   2.17
9     2.45     2.31  2.21   2.17
10    2.45     2.30  2.19   2.15
11    2.45     2.30  2.20   2.16
12    2.44     2.29  2.19   2.16
13    2.43     2.28  2.18   2.15
14    2.43     2.28  2.18   2.15
}\suppCurvesGfid

\def\suppCurvesRefRfid{0.73}
\def\suppCurvesRefGfid{2.50}

\pgfplotsset{
  supp shiftlog/.style={
    y coord trafo/.code={\pgfmathparse{ln(max(##1-#1,1e-4))}},
    y coord inv trafo/.code={\pgfmathparse{exp(##1)+#1}}},
  supp rfid scale/.style={supp shiftlog=0.32, ymin=0.335, ymax=1.75,
    ytick={0.35,0.40,0.50,0.75,1.50}, yticklabels={0.35,0.40,0.50,0.75,1.50}},
  supp gfid scale/.style={supp shiftlog=2.08, ymin=2.135, ymax=3.72,
    ytick={2.2,2.3,2.5,3.5}, yticklabels={2.2,2.3,2.5,3.5}},
}

\definecolor{suppLat}{HTML}{7A7A7A}
\definecolor{suppCls}{HTML}{D55E00}
\definecolor{suppPatch}{HTML}{0072B2}
\definecolor{suppBoth}{HTML}{009E73}

\pgfplotsset{
  supp curves/.style={
    scale only axis, width=0.5\linewidth-1.2cm, height=2.5cm,
    xmin=0.6, xmax=14.4, xtick={2,4,...,14},
    xlabel={Epoch},
    tick label style={font=\scriptsize},
    label style={font=\scriptsize},
    xlabel shift=-4pt, ylabel shift=-3pt,
    tick align=outside, tick pos=left,
    major tick length=2pt,
    axis line style={black!60}, tick style={black!60},
    ymajorgrids, grid style={black!12, line width=0.3pt},
    table/x=epoch,
    unbounded coords=jump,
  },
  supp lat/.style={suppLat,   line width=0.7pt, mark=*,         mark size=0.9pt},
  supp cls/.style={suppCls,   line width=0.7pt, mark=triangle*, mark size=1.2pt},
  supp patch/.style={suppPatch, line width=0.7pt, mark=square*,   mark size=0.8pt},
  supp both/.style={suppBoth,  line width=0.7pt, mark=diamond*,  mark size=1.3pt},
  supp ref/.style={black, dashed, line width=0.7pt, mark=none},
}

\def\suppCurvesPlots#1#2{%
  \ifx#2\empty\else
    \addplot[supp ref, forget plot, domain=0.6:14.4, samples=2] {#2};
  \fi
  \addplot[supp lat,   forget plot] table[y=latents]  {#1};
  \addplot[supp cls,   forget plot] table[y=cls]      {#1};
  \addplot[supp patch, forget plot] table[y=patch]    {#1};
  \addplot[supp both,  forget plot] table[y=clspatch] {#1};
}

\begin{figure}[t]
  \centering
  \pgfplotslegendfromname{suppCurvesLegend}\par\smallskip
  \begin{tikzpicture}
      \begin{axis}[supp curves, supp rfid scale, ylabel={\rfid},
          legend to name=suppCurvesLegend, legend columns=3, legend cell align=left,
          legend style={draw=none, font=\scriptsize,
            /tikz/every even column/.append style={column sep=6pt}, row sep=-1pt},
          legend image post style={xscale=0.7}]
        \addlegendimage{supp lat}   \addlegendentry{Latents only}
        \addlegendimage{supp cls}   \addlegendentry{\cls}
        \addlegendimage{supp patch} \addlegendentry{Patch}
        \addlegendimage{supp both}  \addlegendentry{{\cls}~+~patch}
        \addlegendimage{supp ref}   \addlegendentry{\sdvaeftema}
        \suppCurvesPlots{\suppCurvesRfid}{\suppCurvesRefRfid}
      \end{axis}
  \end{tikzpicture}\hfill
  \begin{tikzpicture}
      \begin{axis}[supp curves, supp gfid scale, ylabel={\gfid}]
        \suppCurvesPlots{\suppCurvesGfid}{\suppCurvesRefGfid}
      \end{axis}
  \end{tikzpicture}
  \caption{\textbf{Decoder training curves.} \rfid (left) and \gfid on \texttt{REGLUE-XL/2} (right) over training, for the latents-only decoder and the semantic-aware decoders conditioned on \cls, patch-level, or both semantics. Dashed line: \sdvaeftema decoder.}
  \label{fig:supp_curves}
\end{figure}
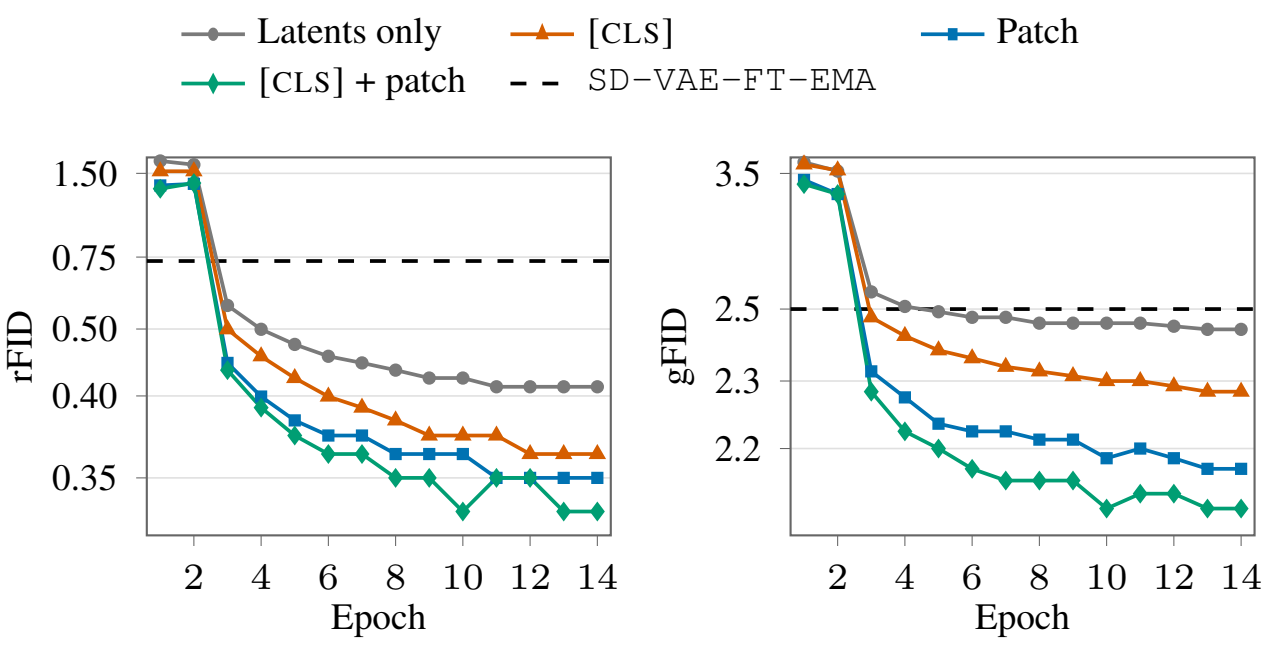

\subsection{Additional Metrics}
\label{supp:metrics}

We report PSNR, SSIM, and LPIPS alongside \rfid for reconstruction, and sFID, Inception Score (IS), Precision, and Recall alongside \gfid for generation.

\parag{Reconstruction.}
In \tableref{tab:supp_recon}, row~(g) uses patch features and the \cls token encoded from the original image, achieving $0.34$ \rfid compared with $0.41$ for the latents-only \vitb. The proxy rows use VFM features extracted from a latents-only reconstruction, while the projected rows use features predicted directly from $\rvz_r$. With either compressed or uncompressed features, both approaches improve \rfid, PSNR, and SSIM over the latents-only \vitb and match or improve LPIPS.

\parag{Jointly generated semantics.}
At both \texttt{B/2} and \texttt{XL/2} scales, conditioning on jointly generated semantics improves \gfid, sFID, and IS over the latents-only transformer (\tableref{tab:supp_gen_b} and \tableref{tab:supp_gen_xl_generated}\hspace{-3pt}~\textcolor{cvprblue}{(a)}). Precision matches or exceeds the baseline.

\parag{Proxy and projected semantics.}
On \sitxl and \repaxl, which generate only image latents, decoding with uncompressed projected semantics (Stage~2) improves \gfid, sFID, IS, and Precision over the latents-only transformer (\tableref{tab:supp_gen_xl_generated}\hspace{-3pt}~\textcolor{cvprblue}{(b)}). Across all \texttt{XL/2} settings in \tableref{tab:supp_gen_xl_proxy_proj}, proxy semantics improve \gfid and IS over latents-only decoding, with equal or slightly higher sFID. For both compressed and uncompressed features, Stage~2 projected semantics achieve lower \gfid and sFID than either proxy semantics or latents-only decoding.


\begin{table*}[t]
\centering
\caption{\textbf{Reconstruction, full metrics.} \imagenet validation set ($50$K images, $256\times256$). The \emph{Semantic-aware}, \emph{Proxy}, and \emph{Projected} rows use the same \vitb backbone as row~(d).
Row~(c) and the rows labeled \emph{Semantic-aware} use features encoded from the original image.
Proxy features are extracted from a latents-only reconstruction, and projected features are predicted directly from $\rvz_r$ after Stage~2. Letters refer to \tableref{tab:decoders_base}.}
\label{tab:supp_recon}
\small
\setlength{\tabcolsep}{6pt}
\begin{tabular}{@{}llcccc@{}}
\toprule
Decoder & Semantics & \rfid$\downarrow$ & PSNR$\uparrow$ & SSIM$\uparrow$ & LPIPS$\downarrow$ \\
\midrule
\textbf{(a)} CNN               & --           & 0.73 &  26.263 & 0.728 & 0.134 \\
\textbf{(b)} CNN, fine-tuned   & --           & 0.66 & 26.221 & 0.733 & 0.136 \\
\textbf{(c)} CNN, fine-tuned   & patch        & 0.60 & 26.194 & 0.730 & 0.136 \\
\midrule
\textbf{(d)} Latents-only \vitb & --          & 0.41 & 26.531 & 0.795 & 0.123 \\
\phantom{\textbf{(d)}} Latents-only \vitl & --      & 0.38 & 26.753 & 0.819 & 0.116 \\
\midrule
\textbf{(e)} Semantic-aware    & \cls (\reglue)        & 0.36 & 26.565 & 0.804 & 0.122 \\
\textbf{(f)} Semantic-aware    & patch (\reglue)       & 0.35 & 26.500 & 0.799 & 0.124 \\
\textbf{(g)} Semantic-aware    & patch + \cls (\reglue)& 0.34 & 26.506 & 0.806 & 0.124 \\
\textbf{(j)} Semantic-aware & \cls (\reg) & 0.34 & 26.553 & 0.805 & 0.122 \\
\phantom{\textbf{(g)}} Semantic-aware & PCA-compressed patches (\redi) & 0.37 & 26.557 & 0.802 & 0.122 \\
\midrule
\phantom{\textbf{(g)}} Proxy             & compressed (\reglue) & 0.37 & 26.548 & 0.808 & 0.122  \\
\phantom{\textbf{(g)}} Projected & compressed (\reglue) & 0.36 & 26.582 & 0.805 & 0.121 \\
\phantom{\textbf{(g)}} Proxy             & uncompressed (\dinovtwob) & 0.38 & 26.536 & 0.801 & 0.123 \\
\phantom{\textbf{(g)}} Projected & uncompressed (\dinovtwob)& 0.35 & 26.597 & 0.811 & 0.121 \\
\bottomrule
\end{tabular}
\end{table*}

\begin{table}[t]
\centering
\caption{\textbf{Semantic-aware decoding on \texttt{B/2} generators.}
For each generator, all decoders use the same $50$K latents sampled from its $400$K-iteration checkpoint without classifier-free guidance.
The indicated semantic features are generated jointly with the latents. The CNN baseline is \sdvaeftema. Rows~(b) and~(c) are fine-tuned from the original \sdvae decoder. Letters refer to~\tableref{tab:decoders_base}.}
\label{tab:supp_gen_b}
\small
\setlength{\tabcolsep}{1pt}
\resizebox{\linewidth}{!}{%
\begin{tabular}{@{}llccccc@{}}
\toprule
Generator & Decoder & \gfid$\downarrow$ & sFID$\downarrow$ & IS$\uparrow$ & Pre.$\uparrow$ & Rec.$\uparrow$ \\
\midrule
\multirow{7}{*}{\texttt{REGLUE-B/2}}
 & \textbf{(a)} CNN                    & 12.90 & 5.77 & 91.1 & 0.67 & 0.67 \\
 & \textbf{(b)} CNN, fine-tuned        & 13.29 & 5.81 & 90.0 &  0.67 & 0.62 \\
 & \textbf{(c)} CNN, fine-tuned + patch & 12.05 & 5.71 & 93.7 & 0.70 & 0.62 \\
 & \textbf{(d)} Latents-only           & 12.08 & 6.20 & 92.7 & 0.70 & 0.61 \\
 & \textbf{(e)} + \cls                 & 10.91 & 6.14 & 98.4 & 0.71 & 0.60 \\
 & \textbf{(f)} + patch                & 10.42 & 6.00 & 100.2 & 0.72 & 0.60 \\
 & \textbf{(g)} + patch + \cls         & 10.13 & 5.95 & 101.5 & 0.72 & 0.60 \\
\midrule
\multirow{3}{*}{\texttt{REG-B/2}}
 & \textbf{(h)} CNN                                 & 15.22 & 6.69 & 94.6 & 0.64 & 0.63 \\
 & \textbf{(i)} Latents-only                        & 14.15 & 6.77 & 95.8 & 0.67 & 0.61 \\
 & \textbf{(j)} + \cls                              & 12.28 & 6.63 & 105.1 & 0.69 & 0.61 \\
\midrule
\multirow{5}{*}{\begin{tabular}[c]{@{}l@{}}\texttt{REGLUE-B/2}\\(unconditional)\end{tabular}}
 & CNN                                 & 27.95 & 6.47 & 39.4 & 0.59 & 0.64 \\
 & Latents-only                        & 26.25 & 6.78 & 40.1 & 0.61 & 0.63 \\
 & + \cls                              & 24.79 & 6.74 & 41.9 & 0.62 & 0.62 \\
 & + patch                             & 24.05 & 6.56 & 42.1 & 0.63 & 0.62 \\
 & + patch + \cls                      & 23.68 & 6.45 & 42.6 & 0.63 & 0.62 \\
\bottomrule
\end{tabular}}
\end{table}

\begin{table}[t]
\centering
\caption{\textbf{Semantic-aware decoding on \texttt{XL/2} generators.}
Full metrics on $50$K samples per setting without classifier-free guidance. CNN denotes \sdvaeftema.
In (a), the decoder uses jointly generated patch features and the \cls token for \reglue, the \cls token for \reg, and $8$-channel patch features for \redi and \coredi.
In (b), it uses uncompressed \dinovtwob patch features and the \cls token predicted from $\rvz_r$ by the Stage~2 projector.}
\label{tab:supp_gen_xl_generated}
\small
\setlength{\tabcolsep}{2.2pt}
\begin{tabular}{@{}llccccc@{}}
\toprule
Generator & Decoder & \gfid$\downarrow$ & sFID$\downarrow$ & IS$\uparrow$ & Pre.$\uparrow$ & Rec.$\uparrow$ \\
\midrule
\multicolumn{7}{@{}l}{\textit{(a) Joint generative models}} \\
\addlinespace[1pt]
\multirow{3}{*}{\regluexl}
 & CNN            & 2.50 & 4.10 &  188.6 & 0.76 & 0.65 \\
 & Latents-only   & 2.43 & 4.07 & 189.4 & 0.79 & 0.63 \\
 & + generated    & 2.15 & 4.05 & 199.5 & 0.79 & 0.62 \\
\midrule
\multirow{3}{*}{\regxl}
 & CNN            & 1.80 &  4.59 & 230.8 & 0.77 & 0.66 \\
 & Latents-only   & 1.63 & 4.36 & 235.6 & 0.79 & 0.66 \\
 & + generated    & 1.54 & 4.32 & 244.4 & 0.80 & 0.64 \\
\midrule
\multirow{3}{*}{\redixl}
 & CNN            & 3.30 &  4.80 & 188.9 &  0.74 &  0.68 \\
 & Latents-only   & 2.85 & 4.74 & 191.3 & 0.77 & 0.64 \\
 & + generated    & 2.65 & 4.65 & 195.7 & 0.78 & 0.63 \\
\midrule
\multirow{3}{*}{\coredixl}
 & CNN            & 3.30 &  4.40 & 176.8 &  0.74 &  0.66 \\
 & Latents-only   & 3.06 & 4.22 & 177.5 & 0.77 & 0.64 \\
 & + generated    & 2.79 & 4.19 & 182.4 & 0.78 & 0.63 \\
\midrule
\multirow{3}{*}{\begin{tabular}[c]{@{}l@{}}\regluexl\\(uncond.)\end{tabular}}
 & CNN            & 9.47 & 4.67 & 89.0 & 0.67 & 0.68 \\
 & Latents-only   & 8.63 & 4.48 & 89.8 & 0.69 & 0.66 \\
 & + generated    & 7.52 & 4.44 & 95.0 & 0.71 & 0.66 \\
\midrule
\multicolumn{7}{@{}l}{\textit{(b) Standard latent generative models}} \\
\addlinespace[1pt]
\multirow{3}{*}{\sitxl}
 & CNN            & \phantom{$^\dagger$}8.60$^\dagger$ & 6.25 & 130.9  & 0.69 & 0.67 \\
 & Latents-only   & 7.58 & 6.06 & 134.0 & 0.71 & 0.66 \\
 & + projected         & 6.58 & 6.01 & 140.1 & 0.72 & 0.65 \\
\midrule
\multirow{3}{*}{\repaxl}
 & CNN            & 5.90 &  5.73 & 157.8 &  0.70 & 0.69 \\
 & Latents-only   & 4.95 & 5.50 & 160.2 & 0.72 & 0.67 \\
 & + projected         & 4.19 & 5.45 & 167.1 & 0.74 & 0.66 \\
\bottomrule \\
[-8pt]
    \multicolumn{7}{@{}c}{\small{$\dagger$: our reproduction; 8.30 reported in~\citep{ma2024sit}}} \\[-2pt]
\end{tabular}
\end{table}

\begin{table}[t]
\centering
\caption{\textbf{Generation at \texttt{XL/2} scale with proxy and projected semantics, full metrics.} $50$K samples, no guidance. 
Both proxy and projected semantics are derived from the generated latents $\rvz_r$.
Unshaded rows use $16$-channel patch features obtained by applying the \reglue compressor.
\colorbox{uncomp}{Shaded} rows use uncompressed $768$-channel patch features.
Both representations include the \cls token.
CNN and latents-only baselines are in~\tableref{tab:supp_gen_xl_generated}.}
\label{tab:supp_gen_xl_proxy_proj}
\small
\setlength{\tabcolsep}{3pt}
\resizebox{\linewidth}{!}{%
\begin{tabular}{@{}llccccc}
\toprule
Generator & Decoder & \gfid$\downarrow$ & sFID$\downarrow$ & IS$\uparrow$ & Pre.$\uparrow$ & Rec.$\uparrow$ \\
\midrule
 & + proxy             & 2.32 & 4.10 & 193.4 & 0.78 & 0.63 \\
 & + projected (St. 1) & 2.31 & 4.12 & 194.2 & 0.79 & 0.63 \\
 & + projected (St. 2)        & 2.26 & 4.05 & 194.6 & 0.79 & 0.62 \\
\uc & + proxy          & 2.30 & 4.12 & 193.6  & 0.78  & 0.62  \\
\uc & + projected (St. 1) & 2.38 & 4.36 & 192.9 & 0.78 & 0.62 \\
\uc\multirow{-6}{*}{\regluexl}
    & + projected (St. 2)        & 2.20 & 4.05 & 196.2 & 0.79 & 0.62 \\
\midrule
 & + proxy             & 1.58 & 4.40 & 240.2 & 0.80 & 0.65 \\
 & + projected (St. 1) & 1.60 & 4.41 & 240.4 & 0.80 & 0.64 \\
 & + projected (St. 2)        & 1.55 & 4.32 & 241.1 & 0.80 & 0.64 \\
\uc & + proxy             & 1.56 & 4.42 & 240.2 & 0.79 & 0.65 \\
\uc & + projected (St. 1) & 1.72 & 4.71 & 238.6 & 0.79 & 0.65 \\
\uc\multirow{-6}{*}{\regxl}
    & + projected (St. 2)        & 1.54 & 4.35 & 241.3 & 0.80 & 0.64 \\
\midrule
 & + proxy             & 2.67 & 4.81 & 195.5 & 0.78 & 0.64 \\
 & + projected (St. 1) & 2.58 & 4.75 & 196.7 & 0.78 & 0.63 \\
 & + projected (St. 2)        & 2.52 & 4.69 & 197.5 & 0.78 & 0.63 \\
\uc & + proxy             & 2.71 & 4.87 & 195.0 & 0.77 & 0.64 \\
\uc & + projected (St. 1) & 2.68 & 4.99 & 194.6 & 0.78 & 0.63 \\
\uc\multirow{-6}{*}{\redixl}
    & + projected (St. 2)        & 2.46 & 4.69 & 198.0 & 0.78 & 0.63 \\
\midrule
 & + proxy             & 2.86 & 4.27 & 181.3 & 0.77 & 0.64 \\
 & + projected (St. 1) & 2.78 & 4.24 & 181.9 & 0.78 & 0.64 \\
 & + projected (St. 2)      & 2.72 & 4.19 & 182.6 & 0.78 & 0.63 \\
\uc & + proxy             & 2.87 & 4.31 & 180.9 & 0.77 & 0.64 \\
\uc & + projected (St. 1) & 2.84 & 4.46 & 180.2 & 0.77 & 0.63 \\
\uc\multirow{-6}{*}{\coredixl}
    & + projected (St. 2)       & 2.64 & 4.19 & 183.3 &  0.78 & 0.63 \\
\midrule
 & + proxy             & 6.83 & 6.06 & 154.2 & 0.68 & 0.69 \\
 & + projected (St. 1) & 6.89 & 6.12 & 138.5 & 0.72 & 0.65 \\
 & + projected (St. 2)      & 6.76 & 6.01 & 139.1 & 0.72 & 0.65 \\
\uc & + proxy             & 7.37 & 6.31 & 136.5 & 0.71 & 0.65 \\
\uc & + projected (St. 1) & 6.92 & 6.38 & 136.2 & 0.72 & 0.65 \\
\uc\multirow{-6}{*}{\sitxl}
    & + projected (St. 2)       & 6.58 & 6.01 & 140.1 & 0.72 & 0.65 \\
\midrule
 & + proxy             & 4.63 & 5.64 & 164.4 & 0.73 & 0.66 \\
 & + projected (St. 1) & 4.44 & 5.56 & 165.1 & 0.73 & 0.67 \\
 & + projected  (St. 2)       & 4.32 & 5.45 & 166.6 & 0.74 & 0.66 \\
\uc & + proxy             & 4.70 & 5.67 & 163.6 & 0.73 & 0.67 \\
\uc & + projected (St. 1) & 4.55 & 5.85 & 163.3 & 0.73 & 0.66 \\
\uc\multirow{-6}{*}{\repaxl}
    & + projected (St. 2)        & 4.19 & 5.45 & 167.1 & 0.74 & 0.66 \\
\midrule
 & + proxy             & 8.19 & 4.56 & 92.0 & 0.70 & 0.66 \\
 & + projected (St. 1) & 7.90 & 4.51 & 92.7 & 0.70 & 0.66 \\
 & + projected (St. 2)        & 7.82 & 4.45 & 93.2 & 0.70 & 0.66 \\
\uc & + proxy             & 8.31 & 4.60 & 91.7 & 0.70 & 0.66 \\
\uc & + projected (St. 1) & 7.86 & 4.70 & 91.9 & 0.70 & 0.66 \\
\uc\multirow{-6}{*}{\shortstack[l]{\regluexl\\(uncond.)}}
    & + projected (St. 2)        & 7.64 & 4.43 & 93.8 & 0.71 & 0.66 \\
\bottomrule
\end{tabular}}
\end{table}

\subsection{Fusion Strategies for Patch Semantics}
\label{supp:fusion_strategy}

We compare channel-wise and token-wise fusion of the patch semantics jointly generated by \regluexl. In both variants, we upsample the patch features to the $32\times32$ latent grid and include the global \cls token. Channel-wise fusion concatenates each patch feature with the latent at the same spatial position, yielding $1025$ tokens in total. Token-wise fusion adds the $1024$ patch features as separate tokens, increasing the sequence length to $2049$.

As shown in \tableref{tab:supp_fusion_strategy}, token-wise fusion improves \gfid only slightly, from $2.15$ to $2.13$, while decoder throughput drops by $65.0\%$, from $127.8$ to $44.75$ images/s. We therefore use channel-wise fusion for patch semantics by default.

\begin{table}[h]
\centering
\setlength{\tabcolsep}{4pt}
\small
\caption{\textbf{Fusion of patch semantics in the decoder.}
We compare channel-wise and token-wise concatenation of the patch features and image latents jointly generated by \regluexl.
Both variants include the global \cls token as a separate input token.
We report decoder throughput at inference.}
\label{tab:supp_fusion_strategy}
\begin{tabular}{@{}lccc@{}}
\toprule
Fusion & \#Tokens & Throughput (img/s)$\uparrow$ & gFID$\downarrow$ \\
\midrule
Channel-wise & $1025$      & 127.8\hspace{3.5em} & $2.15$\hspace{3.0em} \\
Token-wise   & $2049$ & $44.75_{\textcolor{red}{\tiny(-65.0\%)}}$ & $2.13_{\textcolor{ForestGreen}{\tiny(-0.9\%)}}$ \\
\bottomrule
\end{tabular}
\end{table}

\subsection{Generation with Classifier-Free Guidance}
\label{supp:cfg}

We also test semantic-aware decoding in the more challenging setting of converged XL/2 generators sampled with classifier-free guidance (CFG). We use the same $250$-step sampler for all four generators. As in the original papers, we use interval guidance~\cite{kynkaanniemi2024applying} for \reglue (160 epochs) and \reg (800 epochs), and apply guidance at every step for \redi (800 epochs) and \coredi (400 epochs). In our experiments, we use $w=2.6$ over $[0.0, 0.85]$ for \reglue and $w=2.4$ over $[0.0, 0.75]$ for \reg. For \redi and \coredi, we use $w=1.9$ and $w=1.7$, respectively. As shown in \tableref{tab:cfg_results}, conditioning the decoder on jointly generated semantics improves \gfid on \reglue ($1.46 \rightarrow 1.44$), \reg ($1.36 \rightarrow 1.32$), and \redi ($1.61 \rightarrow 1.54$), and matches the baseline on \coredi ($1.58$).

\begin{table}[t]
\centering
\small
\caption{\textbf{Classifier-free guidance with semantic-aware decoding.}
We compare each framework's default decoder with our decoder conditioned on jointly generated semantics.
We report \gfid$\downarrow$ on $50$K samples per converged XL/2 generator.}
\label{tab:cfg_results}
\setlength{\tabcolsep}{4pt}
\small
\begin{tabular}{@{}lcccc}
\toprule
 Decoder & \reglue & \reg & \redi & \coredi \\
\midrule
Default          & 1.46 & 1.36 & 1.61 & \textbf{1.58} \\
Semantic-Aware  & \hl\textbf{1.44} & \hl\textbf{1.32} & \hl\textbf{1.54} & \hl\textbf{1.58} \\
\bottomrule
\end{tabular}
\end{table}

\subsection{Decoder Capacity and Efficiency}
\label{supp:additional_decoder_results}

\paragraph{Decoding cost.}
\tableref{tab:supp_cost} reports decoding throughput for $256\times256$ images on a single NVIDIA A100 GPU with a batch size of $64$. On \regluexl, the latents-only \vitb decoder achieves $2.43$ \gfid at $127.8$ images/s. The semantic-aware \vitb decoder matches this throughput and improves \gfid to $2.15$ by conditioning on jointly generated semantics. Scaling the latents-only decoder to \vitl gives $2.41$ \gfid at $46.42$ images/s. We also compare decoding costs when the semantic inputs are derived from the sampled latents themselves. The proxy approach extracts VFM features from a preliminary latents-only reconstruction, while the semantic projector predicts these features directly from the latents. Compared with the proxy approach, decoding with projected semantics achieves $1.9\times$ the throughput, uses about one-third fewer parameters, and improves \gfid on both \regluexl and \sitxl. Note that  diffusion sampling with an \sitxl backbone runs at only $0.56$ images/s, while decoding takes a negligible share of the end-to-end generation time (under $0.5\%$ with the \vitb decoders).

\begin{table}[t]
\centering
\caption{\textbf{Decoding cost.} Per-image cost of each decoding path at $256\times256$ and \gfid$\downarrow$ on two generators. Throughput is measured
on a single A100 with batch size 64.}
\label{tab:supp_cost}
\small
\setlength{\tabcolsep}{3pt}
\small
\begin{tabular}{@{}lccccc@{}}
\toprule
& & & \multicolumn{2}{c}{\gfid$\downarrow$ (XL/2)} \\
\cmidrule(l){4-5}
Decoding path & Params & img/s & \texttt{REGLUE} & \texttt{SiT} \\
\midrule
CNN (\sdvaeftema)          & 49.5M  & 112.1 & 2.50 & 8.60 \\
Latents-only \vitb         & 86.0
M  & 127.8 & 2.43 & 7.58 \\
Latents-only \vitl         & 303.6M & 46.42 & 2.41 & 7.45 \\
Generated semantics        & 86.6M  & 127.8 & 2.15 & \redx   \\
Proxy semantics            & 259.2M & 60.61 & 2.30 & 7.37 \\
Projected semantics        & 173.4M & 116.8 & 2.20 & 6.58 \\
\bottomrule
\end{tabular}
\end{table}

\paragraph{Decoder capacity.}
Since the semantic projector predicts its features from the VAE latents alone, we ask whether a larger latents-only decoder can achieve comparable gains. \figureref{fig:decoder_capacity_curves} compares the latents-only \vitb and \vitl decoders trained for $14$ epochs with the same losses. With $3.5\times$ as many parameters as \vitb, \vitl improves faster in the first few epochs. Both models plateau toward the end of training, and the final advantage of \vitl is only $0.03$ in \rfid and $0.02$ in \gfid on \regluexl. Pairing the semantic projector with a \vitb decoder yields lower \gfid than the latents-only \vitl decoder on each of the six generators (\tableref{tab:decoder_params}). Together, the projector and \vitb decoder use about $43$\% fewer parameters and require $34.7$\% fewer training FLOPs than the latents-only \vitl decoder (\tableref{tab:decoder_compute}). These results suggest that explicitly predicting semantic features makes more effective use of the same latents than scaling the decoder from \vitb to \vitl.

\begin{table}[t]
  \centering
  \caption{\textbf{Training compute.}
  Latents-only \vitl decoder vs.\ our projector followed by a \vitb decoder.
  FLOPs sum all forward and backward passes during training, including those of the frozen encoder, DINOv2, LPIPS and the discriminator; for ours, they cover projector training, decoder training and joint fine-tuning.}
  \setlength{\tabcolsep}{5pt}
  \small
  \begin{tabular}{lcc}
    \toprule
    Model & Params & Training FLOPs \\
    \midrule
    Latents-only \vitl       & 303.6M & $7.2\times10^{19}$\hspace{2.5em} \\
    Projected semantics (ours)   & 173.4M & $4.7\times10^{19}_{\textcolor{ForestGreen}{\tiny(-34.7\%)}}$ \\
    \bottomrule
  \end{tabular}

  \label{tab:decoder_compute}
\end{table}

\pgfplotstableread{
epoch base  large
1     1.69  1.69
2     1.63  0.63
3     0.56  0.48
4     0.50  0.45
5     0.47  0.43
6     0.45  0.42
7     0.44  0.41
8     0.43  0.40
9     0.42  0.40
10    0.42  0.39
11    0.41  0.39
12    0.41  0.38
13    0.41  0.38
14    0.41  0.38
}\capCurvesRfid

\pgfplotstableread{
epoch base  large
1     3.65  3.56
2     3.53  2.74
3     2.57  2.49
4     2.51  2.48
5     2.49  2.47
6     2.47  2.46
7     2.47  2.44
8     2.45  2.43
9     2.45  2.44
10    2.45  2.43
11    2.45  2.43
12    2.44  2.43
13    2.43  2.41
14    2.43  2.41
}\capCurvesGfid

\pgfplotsset{
  cap shiftlog/.style={
    y coord trafo/.code={\pgfmathparse{ln(max(##1-#1,1e-4))}},
    y coord inv trafo/.code={\pgfmathparse{exp(##1)+#1}}},
  cap rfid scale/.style={cap shiftlog=0.30, ymin=0.36, ymax=1.75,
    ytick={0.40,0.50,0.75,1.50}, yticklabels={0.40,0.50,0.75,1.50}},
  cap gfid scale/.style={cap shiftlog=2.30, ymin=2.37, ymax=3.72,
    ytick={2.4,2.5,3.0,3.5}, yticklabels={2.4,2.5,3.0,3.5}},
}

\definecolor{capBase}{HTML}{7A7A7A}
\definecolor{capLarge}{HTML}{CC79A7}

\pgfplotsset{
  cap curves/.style={
    scale only axis, width=0.5\linewidth-1.2cm, height=2.5cm,
    xmin=0.6, xmax=14.4, xtick={2,4,...,14},
    xlabel={Epoch},
    tick label style={font=\scriptsize},
    label style={font=\scriptsize},
    xlabel shift=-4pt, ylabel shift=-3pt,
    tick align=outside, tick pos=left,
    major tick length=2pt,
    axis line style={black!60}, tick style={black!60},
    ymajorgrids, grid style={black!12, line width=0.3pt},
    table/x=epoch,
    unbounded coords=jump,
  },
  cap base/.style={capBase,  line width=0.7pt, mark=*,       mark size=0.9pt},
  cap large/.style={capLarge, line width=0.7pt, mark=square*, mark size=0.8pt},
}

\def\capCurvesPlots#1{%
  \addplot[cap base,  forget plot] table[y=base]  {#1};
  \addplot[cap large, forget plot] table[y=large] {#1};
}

\begin{figure}[t]
  \centering
  \pgfplotslegendfromname{capCurvesLegend}\par\smallskip
  \begin{tikzpicture}
      \begin{axis}[cap curves, cap rfid scale, ylabel={\rfid},
          legend to name=capCurvesLegend, legend columns=2, legend cell align=left,
          legend style={draw=none, font=\scriptsize,
            /tikz/every even column/.append style={column sep=6pt}, row sep=-1pt},
          legend image post style={xscale=0.7}]
        \addlegendimage{cap base}  \addlegendentry{Latents-only \vitb}
        \addlegendimage{cap large} \addlegendentry{Latents-only \vitl}
        \capCurvesPlots{\capCurvesRfid}
      \end{axis}
  \end{tikzpicture}\hfill
  \begin{tikzpicture}
      \begin{axis}[cap curves, cap gfid scale, ylabel={\gfid}]
        \capCurvesPlots{\capCurvesGfid}
      \end{axis}
  \end{tikzpicture}
  \caption{\textbf{Training curves of latents-only decoders.} \rfid (left) and \gfid on \texttt{REGLUE-XL/2} (right) over training for the \vitb and \vitl latents-only decoders. Despite having more than $3\times$ as many parameters, \vitl closely tracks \vitb, and both curves have converged.}
  \label{fig:decoder_capacity_curves}
\end{figure}
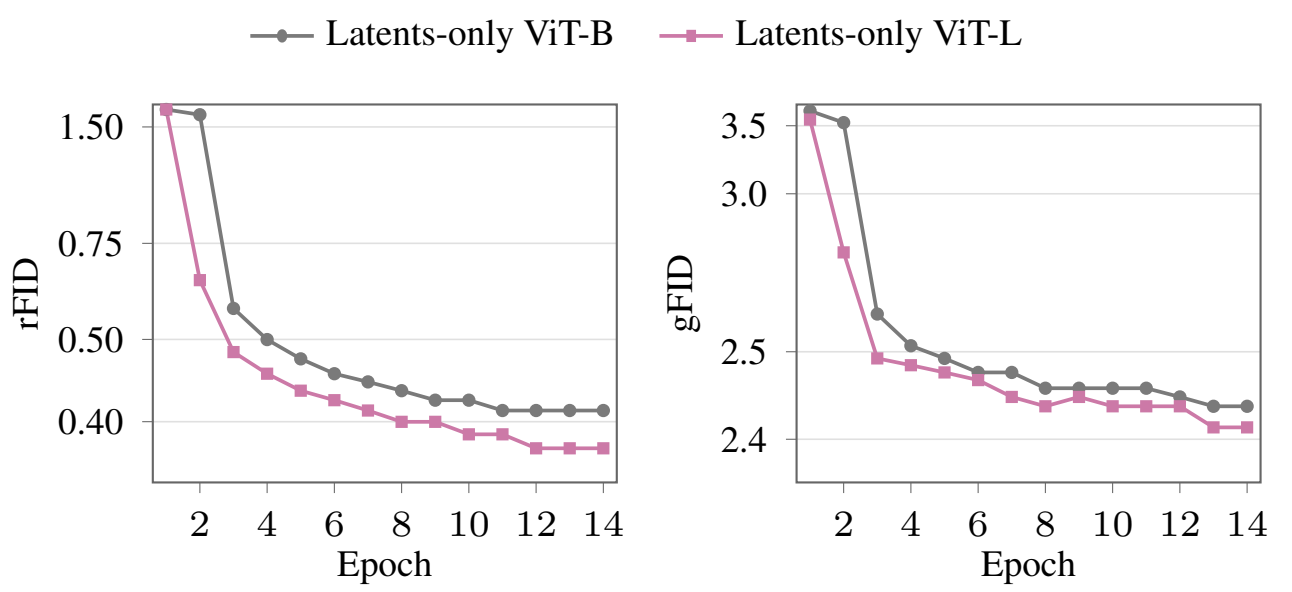

\definecolor{semRedi}{HTML}{B3B3B3}   
\definecolor{semCoredi}{HTML}{EB6834}
\definecolor{semReglue}{HTML}{1BAF7A}
\definecolor{semDino}{HTML}{6666FF}   

\subsection{Semantic Richness of Compressed Features}
\label{supp:additional_results_richness}

On \redi and \coredi, projected semantics outperform the jointly generated ones (\tableref{tab:sem_sources_reglue_reg}).
We argue that this gap stems from semantic richness, and measure how much of \dinovtwob each semantic representation retains.
We compare \redi (PCA, $8$ channels), \coredi (learned linear projection, $8$ channels) and \reglue (nonlinear compressor, $16$ channels) with the uncompressed \dinovtwob patch features and \cls token approximated by our projector.
The \reglue code is also the space of the compressed semantics (\tableref{tab:compression} and \tableref{tab:supp_gen_xl_proxy_proj}).

\parag{Metrics.}
All metrics are computed on frozen \dinovtwob features of real ImageNet images.
\emph{Patch $R^2$} is the fraction of variance explained when a small MLP predicts the original \dinovtwob patch tokens from each compressed representation.
\emph{mkNN} is the fraction of an image's $10$ nearest neighbours in the \dinovtwob \cls-token space that remain among its $10$ nearest neighbours in the compressed space, averaged over images.
\emph{ImageNet attentive probing} \cite{psomas2025attention} and \emph{Cityscapes mIoU} follow~\citep{petsangourakis2026reglue}.
By definition, \dinovtwob scores $1$ on Patch $R^2$ and mkNN, so we omit it from \figureref{fig:supp_info}.

\parag{Results.}
The $8$-channel compressed space of \redi and \coredi keep only a small part of \dinovtwob (\figureref{fig:supp_info}).
They recover $17\%$ and $21\%$ of the patch-token variance and share only $5\%$ of their nearest neighbours with \dinovtwob.
Their probing accuracy is $30.3\%$ and $31.9\%$, against $84.2\%$ for \dinovtwob (\figureref{fig:supp_sem}).
Segmentation degrades less, but still trails \dinovtwob by $13.4$ and $7.5$ mIoU.
On the other hand, the $16$-channel \reglue representation scores higher than its $8$-channel counterparts on all four metrics, indicating that it retains more semantic information.


\parag{Connection to generation.}
We then check whether this ordering carries over to generation by decoding the same \redixl and \coredixl latents with each of the three representations: the $8$-channel, \reglue and \dinovtwob features.
\gfid on \redixl and \coredixl follows the same ordering (\tableref{tab:supp_gen_xl_generated}, \tableref{tab:supp_gen_xl_proxy_proj}).
With their own jointly generated $8$-channel semantics, the decoder reaches $2.65$ on \redi and $2.79$ on \coredi.
Stage~2 projected semantics in the $16$-channel \reglue space lower \gfid to $2.52$ and $2.72$.
Uncompressed projected \dinovtwob features lower it further, to $2.46$ and $2.64$ respectively.
So the richer the semantics given to the decoder, the lower the \gfid.

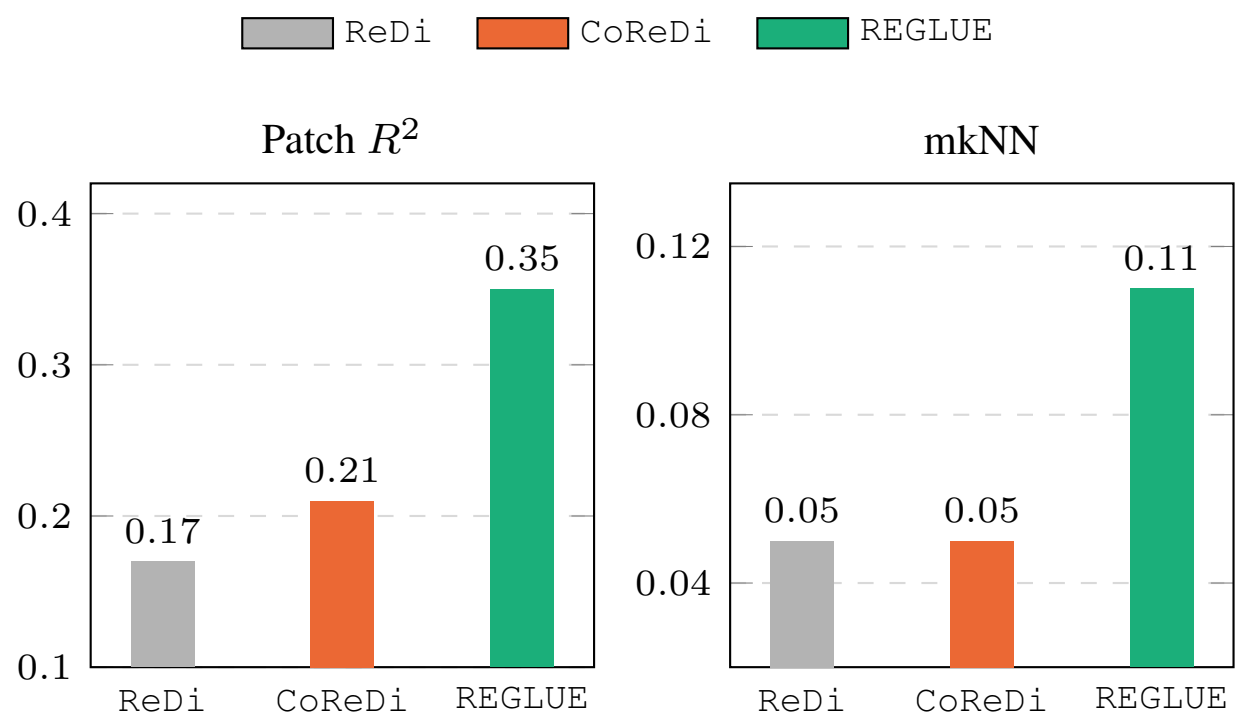
\begin{figure}[t]
  \centering
  \begin{tikzpicture}
    \begin{axis}[
      name=r2ax,
      scale only axis,
      width=0.39\columnwidth,
      height=3.2cm,
      enlarge x limits=0.2,
      symbolic x coords={redi,coredi,reglue},
      xtick={redi,coredi,reglue},
      xticklabels={\redi, \coredi, \reglue},
      xticklabel style={font=\scriptsize},
      major x tick style={draw=none},
      ymin=0.1, ymax=0.42,
      ytick={0,0.1,0.2,0.3,0.4},
      scaled y ticks=false,
      yticklabel style={font=\scriptsize, /pgf/number format/fixed},
      ymajorgrids,
      grid style={dashed, gray!30},
      title={Patch $R^2$},
      title style={font=\footnotesize, yshift=-4pt},
      nodes near coords,
      nodes near coords style={font=\scriptsize, /pgf/number format/fixed,
        /pgf/number format/fixed zerofill, /pgf/number format/precision=2},
      every axis plot/.append style={ybar, bar width=12pt, draw=none},
      area legend,
      legend to name=leg:supp_info,
      legend columns=-1,
      legend style={draw=none, font=\scriptsize,
        /tikz/every even column/.append style={column sep=6pt}},
    ]
      \addplot[fill=semRedi]   coordinates {(redi,0.17)};
      \addplot[fill=semCoredi] coordinates {(coredi,0.21)};
      \addplot[fill=semReglue] coordinates {(reglue,0.35)};
      \legend{{\redi}, {\coredi}, {\reglue}}
    \end{axis}
    \begin{axis}[
      at={(r2ax.south east)}, xshift=0.9cm,
      scale only axis,
      width=0.39\columnwidth,
      height=3.2cm,
      enlarge x limits=0.2,
      symbolic x coords={redi,coredi,reglue},
      xtick={redi,coredi,reglue},
      xticklabels={\redi, \coredi, \reglue},
      xticklabel style={font=\scriptsize},
      major x tick style={draw=none},
      ymin=0.02, ymax=0.135,
      ytick={0,0.04,0.08,0.12},
      scaled y ticks=false,
      yticklabel style={font=\scriptsize, /pgf/number format/fixed},
      ymajorgrids,
      grid style={dashed, gray!30},
      title={mkNN},
      title style={font=\footnotesize, yshift=-4pt},
      nodes near coords,
      nodes near coords style={font=\scriptsize, /pgf/number format/fixed,
        /pgf/number format/fixed zerofill, /pgf/number format/precision=2},
      every axis plot/.append style={ybar, bar width=12pt, draw=none},
    ]
      \addplot[fill=semRedi]   coordinates {(redi,0.05)};
      \addplot[fill=semCoredi] coordinates {(coredi,0.05)};
      \addplot[fill=semReglue] coordinates {(reglue,0.11)};
    \end{axis}
    \node[anchor=south] at ([yshift=2pt]current bounding box.north)
      {\pgfplotslegendfromname{leg:supp_info}};
  \end{tikzpicture}
  \vspace{-0.5em}
  \caption{\textbf{Information kept from \dinovtwob.}
  \dinovtwob scores $1$ on both metrics by definition and is omitted.}
  \label{fig:supp_info}
\end{figure}

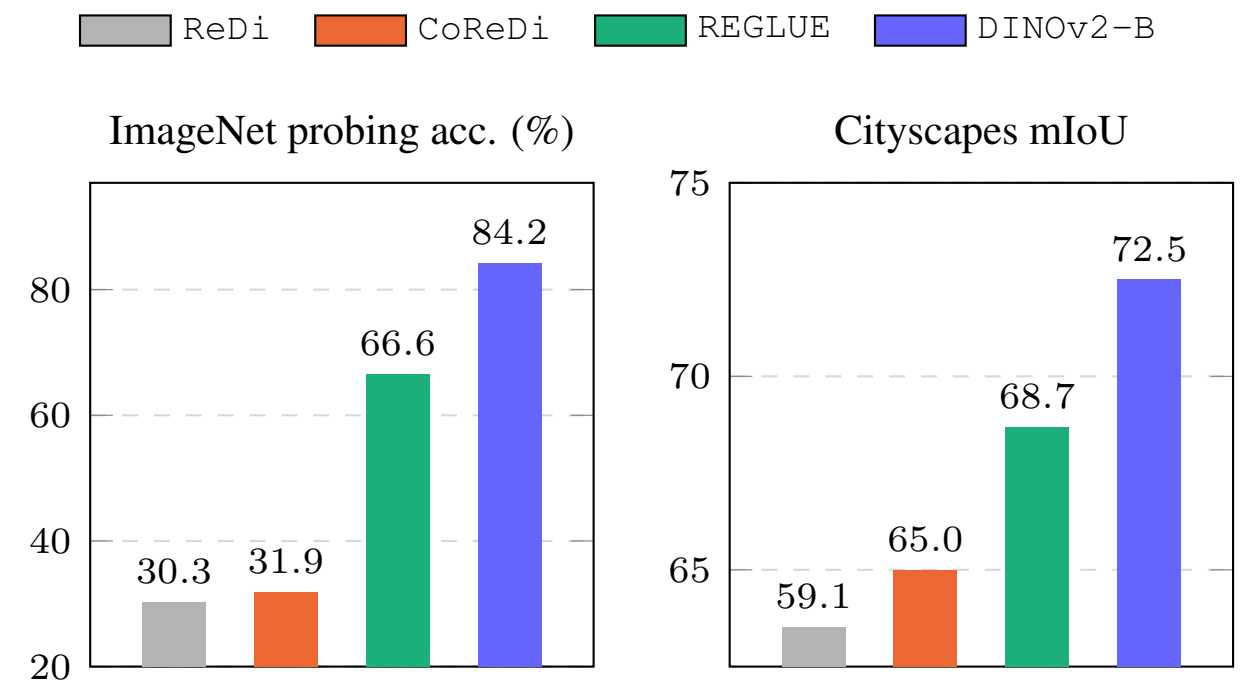
\begin{figure}[t]
  \centering
  \begin{tikzpicture}
    \begin{axis}[
      name=epax,
      scale only axis,
      width=0.39\columnwidth,
      height=3.2cm,
      enlarge x limits=0.25,
      symbolic x coords={redi,coredi,reglue,dino},
      xtick=\empty,
      ymin=20, ymax=97,
      ytick={0,20,40,60,80},
      yticklabel style={font=\scriptsize},
      ymajorgrids,
      grid style={dashed, gray!30},
      title={ImageNet
probing acc. (\%)},
      title style={font=\footnotesize, yshift=-4pt},
      nodes near coords,
      nodes near coords style={font=\scriptsize, /pgf/number format/fixed,
        /pgf/number format/fixed zerofill, /pgf/number format/precision=1},
      every axis plot/.append style={ybar, bar width=12pt, draw=none},
      area legend,
      legend to name=leg:supp_sem,
      legend columns=-1,
      legend style={draw=none, font=\scriptsize,
        /tikz/every even column/.append style={column sep=6pt}},
    ]
      \addplot[fill=semRedi]   coordinates {(redi,30.3)};
      \addplot[fill=semCoredi] coordinates {(coredi,31.9)};
      \addplot[fill=semReglue] coordinates {(reglue,66.6)};
      \addplot[fill=semDino]   coordinates {(dino,84.2)};
      \legend{{\redi}, {\coredi}, {\reglue}, {\dinovtwob}}
    \end{axis}
    \begin{axis}[
      at={(epax.south east)}, xshift=0.9cm,
      scale only axis,
      width=0.39\columnwidth,
      height=3.2cm,
      enlarge x limits=0.25,
      symbolic x coords={redi,coredi,reglue,dino},
      xtick=\empty,
      y coord trafo/.code={\pgfmathparse{#1>=65 ? #1 : 65-(65-#1)*0.25}},
      y coord inv trafo/.code={\pgfmathparse{#1>=65 ? #1 : 65-(65-#1)*4}},
      ymin=55, ymax=75,
      ytick={65,70,75},
      yticklabel style={font=\scriptsize},
      ymajorgrids,
      grid style={dashed, gray!30},
      title={Cityscapes mIoU},
      title style={font=\footnotesize, yshift=-4pt},
      point meta=rawy,
      nodes near coords,
      nodes near coords style={font=\scriptsize, /pgf/number format/fixed,
        /pgf/number format/fixed zerofill, /pgf/number format/precision=1},
      every axis plot/.append style={ybar, bar width=12pt, draw=none},
    ]
      \addplot[fill=semRedi]   coordinates {(redi,59.1)};
      \addplot[fill=semCoredi] coordinates {(coredi,65.0)};
      \addplot[fill=semReglue] coordinates {(reglue,68.7)};
      \addplot[fill=semDino]   coordinates {(dino,72.5)};
    \end{axis}
    \node[anchor=south] at ([yshift=2pt]current bounding box.north)
      {\pgfplotslegendfromname{leg:supp_sem}};
  \end{tikzpicture}
  \vspace{-0.5em}
  \caption{\textbf{Semantic quality.}
  ImageNet Attentive Probing accuracy and Cityscapes mIoU.}
  \label{fig:supp_sem}
\end{figure}


\subsection{SD-VAE Decoder Baselines}
\label{sec:supp_original_sdvae}

We fine-tune our CNN decoders (\tableref{tab:decoders_base}) from the original \sdvae decoder~\cite{rombach2022high}.
The commonly used \sdvaeftema checkpoint~\citep{sdvaeftema} was also fine-tuned from this decoder, using additional images from LAION-Aesthetics and LAION-Humans while preserving the original latent space.
It achieves lower \gfid than the original decoder on every generator in \tableref{tab:supp_original_sdvae}. We therefore use \sdvaeftema as the stronger baseline for our CNN comparisons.

\begin{table}[t]
\centering
\caption{\textbf{Original  vs.\ SD-VAE-FT-EMA decoder.} gFID$\downarrow$ without classifier-free guidance. Both CNN decoders share the frozen SD-VAE encoder and decode the same latents $\mathbf{z}_r$. $^\dagger$: our reproduction.}
\label{tab:supp_original_sdvae}
\small
\setlength{\tabcolsep}{2.8pt}
\begin{tabular}{l cc}
\toprule
Generator & Original \sdvae & \sdvaeftema \\
\midrule

\regluebase & 14.27 & 12.90 \\
\regbase    & 17.16 & 15.22 \\
\midrule
\coredixl &  4.09 &  3.30 \\
\redixl   &  4.00 &  3.30 \\
\regluexl &  3.22 &  2.50 \\
\regxl    &  2.26 &  1.80 \\
\sitxl    &  9.73 &  \phantom{$^\dagger$}8.60$^\dagger$ \\
\repaxl   &  6.80 &  5.90 \\
\midrule
\regluebase uncond.    & 29.11 & 27.95 \\
\regluexl uncond.   & 10.52 &  9.47 \\
\bottomrule
\end{tabular}
\end{table}

\section{Additional Implementation Details}
\label{supp:setup}

\subsection{Input Preprocessing}
For each training image, we take one random square crop and resize it to $256\times256$ for the SD-VAE encoder and to $224\times224$ for DINOv2-B (ImageNet normalization). Both grids then cover the same region: each DINOv2 patch of the $16\times16$ grid spans a $2\times2$ block of the $32\times32$ latent grid. In reconstruction evaluation, the encoder receives a $256\times256$ center crop.
Since generators are trained on SD-VAE latents multiplied by $0.18215$, we divide their samples by this factor before decoding, as our decoders are trained on the raw encoder outputs. 

\subsection{Decoder Architectures and Training}

\tableref{tab:supp_hparams} summarizes the decoder architectures and training settings.
All decoder variants use $\mathcal{L}_{\text{dec}}$ from \equationref{eq:dec}, which combines an $\ell_1$ reconstruction loss, the LPIPS perceptual loss, and an adaptively weighted adversarial loss.
We train the transformer decoders, including the semantic-aware decoder $\gD_{\text{sem}}$ (\equationref{eq}), from scratch for $14$ epochs.
We fine-tune both CNN variants from the original \sdvae decoder for $5$ epochs. The semantic-aware CNN adds a zero-initialized input convolution for patch features. We train each decoder using 4$\times$A100 GPUs.
\begin{table*}[t]
\centering
\caption{\textbf{Architecture and training hyperparameters of the decoders and semantic projector.}
For the semantic-aware transformer decoder $\gD_{\text{sem}}$, only the input width varies across semantic representations.
The input comprises four latent channels and $C_s$ patch-feature channels, with $C_s=16$ for \reglue, $8$ for \redi and \coredi, $768$ for uncompressed \dinovtwo features, and $0$ for \cls-only conditioning.
$^\dagger$\,The parameter count depends on $C_s$.}
\label{tab:supp_hparams}
\small
\setlength{\tabcolsep}{3pt}
\resizebox{\textwidth}{!}{%
\begin{tabular}{@{}lcccccc@{}}
\toprule
& \multicolumn{3}{c}{Transformer decoders} & & \multicolumn{2}{c}{Semantic projector} \\
\cmidrule(lr){2-4}\cmidrule(l){6-7}
& Latents-only & Latents-only & Semantic-aware $\gD_{\text{sem}}$ & CNN & Stage~1 & Stage~2 \\
\midrule
\multicolumn{7}{@{}l}{\textit{Architecture}} \\
Backbone              & \vitb & \vitl & \vitb & \sdvae decoder & \vitb ($\gP$) & $\gP$ + $\gD_{\text{sem}}$ \\
Depth / width / heads & 12 / 768 / 12 & 24 / 1024 / 16 & 12 / 768 / 12 & -- & 12 / 768 / 12 & -- \\
Input channels        & 4 & 4 & $4+C_s$ & $4+16$ (zero-init) & 4 & -- \\
Parameters            & 86.0M & 303.6M & 86--87M$^\dagger$ & 49.5M & 86.8M & 173.4M \\
\midrule
\multicolumn{7}{@{}l}{\textit{Optimization}} \\
Initialization        & random & random & random & \sdvae from~\cite{rombach2022high} & random & Stage~1 + 14-ep.\ $D_\text{sem}$ \\
Epochs                & 14 & 14 & 14 & 5 & 20 & 2 \\
Batch size            & 96 & 64 & 96 & 96 & 512 & 96 \\
Optimizer             & Adam & Adam & Adam & Adam & Adam & Adam \\
$(\beta_1,\beta_2)$             & (0.5, 0.9) & (0.5, 0.9) & (0.5, 0.9) & (0.5, 0.9) & (0.9, 0.999) & (0.5, 0.9) \\
Learning rate         & $4.32\times10^{-4}$ & $2.88\times10^{-4}$ & $4.32\times10^{-4}$ & $4.32\times10^{-4}$ & $1\times10^{-4}$ & $4.32\times10^{-4}$ \\
Schedule / warm-up    & Constant / None & Constant / None & Constant / None & Constant / None & Constant / None & Constant / None \\
Weight decay / EMA    & 0 / 0.9999 & 0 / 0.9999 & 0 / 0.9999 &0 / 0.9999 & 0 / none & 0 / 0.9999 \\
\midrule
\multicolumn{7}{@{}l}{\textit{Objective}} \\
Losses                & $\mathcal{L}_{\text{dec}}$ & $\mathcal{L}_{\text{dec}}$ & $\mathcal{L}_{\text{dec}}$ & $\mathcal{L}_{\text{dec}}$ & $\mathcal{L}_{\text{proj}}$ & $\mathcal{L}_{\text{dec}}+\mathcal{L}_{\text{proj}}$ \\
$\lambda_{\text{perc}}$ & 1.0 & 1.0 & 1.0 & 1.0 & -- & 1.0 \\
${\lambda_{\text{adv}}}^{*}$ & $0.75 \cdot \frac{\lVert\nabla\mathcal{L}_{\text{rec}}\rVert}{\lVert\nabla\mathcal{L}_{\text{GAN}}\rVert + \epsilon}$ & $0.75 \cdot \frac{\lVert\nabla\mathcal{L}_{\text{rec}}\rVert}{\lVert\nabla\mathcal{L}_{\text{GAN}}\rVert + \epsilon}$ & $0.75 \cdot \frac{\lVert\nabla\mathcal{L}_{\text{rec}}\rVert}{\lVert\nabla\mathcal{L}_{\text{GAN}}\rVert + \epsilon}$ & $0.5 \cdot \frac{\lVert\nabla\mathcal{L}_{\text{rec}}\rVert}{\lVert\nabla\mathcal{L}_{\text{GAN}}\rVert + \epsilon}$ & -- & $0.75 \cdot \frac{\lVert\nabla\mathcal{L}_{\text{rec}}\rVert}{\lVert\nabla\mathcal{L}_{\text{GAN}}\rVert + \epsilon}$ \\
Discriminator start epoch & 2 & 2 & 2 & 0 & -- & 0 \\
Adversarial loss start epoch & 3 & 3 & 3 & 0 & -- & 0 \\
\bottomrule
    \multicolumn{7}{@{}c}{\footnotesize $^{*}$$\epsilon{=}10^{-6}$ for the transformer decoders and Stage~2, $\epsilon{=}10^{-4}$ for the CNN (following~\cite{rombach2022high}).}
    \end{tabular}}
\end{table*}

\subsection{Semantic Projector}
\label{supp:projector}
\tableref{tab:supp_hparams} summarizes the architecture and optimization settings for our \vitb semantic projector $\gP$.
In Stage~1, we train only $\gP$ for $20$ epochs to predict semantic features from $\rvz_r$ using the projection loss $\mathcal{L}_{\text{proj}}$ (\equationref{eq:proj}).
In Stage~2, we jointly fine-tune $\gP$ and $\gD_{\text{sem}}$ for $2$ additional epochs by minimizing $\mathcal{L}_{\text{dec}}+\mathcal{L}_{\text{proj}}$, where $\mathcal{L}_{\text{dec}}$ is the decoder loss defined in \equationref{eq:dec}.

\paragraph{Compressed semantics.}
The projector $\gP$ predicts the \reglue representation $\mathbf{z}_s=(\mathcal{S}(\mathbf{f}_{\text{patch}}),\mathbf{f}_{\text{cls}})$ directly from the \sdvae latent $\mathbf{z}_r$.
To construct the patch targets, we encode the image with the frozen \dinovtwob encoder, concatenate the patch features from its last four blocks along the channel dimension, and apply the frozen \reglue compressor $\mathcal{S}$. This yields a $16\times16$ grid of $16$-channel patch targets. As in \reglue, we keep the encoder's final-layer \cls token uncompressed and use it as the global target for $\gP$.

The \vitb projector embeds non-overlapping $2\times2$ patches of $\mathbf{z}_r$, producing one token per location in the target grid, and prepends a learnable \cls token.
One linear head predicts the compressed features at each location from the corresponding output patch token. A second linear head predicts the global target from the output \cls token. We supervise both predictions with the projection loss in \equationref{eq:proj}.

\paragraph{Uncompressed semantics.}
Here $\mathcal{C}$ is the identity and $\mathcal{P}$ predicts the last-layer \texttt{DINOv2-B} patch features ($768\times16\times16$) and [CLS] token directly.
The architecture and objective are the same as in the compressed setting; only the patch head changes, outputting 768 channels instead of 16.

\section{Visualizations}
\label{sec:supp_vis}

In the main paper, \figureref{fig:qualitative} compares \sdvaeftema with our semantic-aware decoder on samples from \regluexl, which jointly generates the image latents and semantics.
Here, we qualitatively assess how faithfully the same two decoders reconstruct real images.
\figureref{fig:supp_recon} shows original images (left), their reconstructions by the \sdvaeftema decoder (middle), and reconstructions by our semantic-aware decoder (right). Both decoders receive the same \sdvae latents. Ours also receives the image's semantics in the \reglue space. Our decoder better preserves regular high-frequency structures, such as the cube mosaic behind the scooter, the abacus beads, and the basket weave, which \sdvaeftema distorts into irregular patterns. It also keeps colors closer to the original, e.g., the eye spots on the peacock's feathers.

\begin{figure*}[t]
  \centering
  \setlength{\tabcolsep}{1pt}
  \begin{tabular}{ccc}
    \small Original & \small \sdvaeftema & \small Ours \\
    \includegraphics[width=0.24\textwidth]{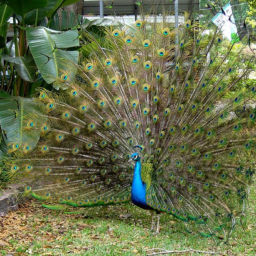} &
    \includegraphics[width=0.24\textwidth]{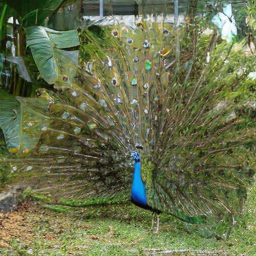} &
    \includegraphics[width=0.24\textwidth]{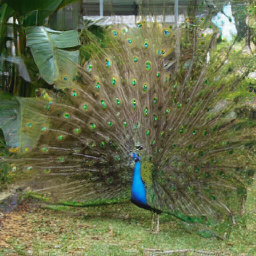} \\
    \includegraphics[width=0.24\textwidth]{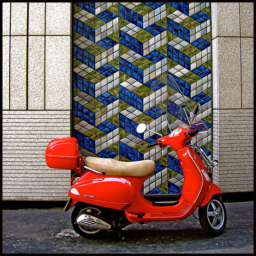} &
    \includegraphics[width=0.24\textwidth]{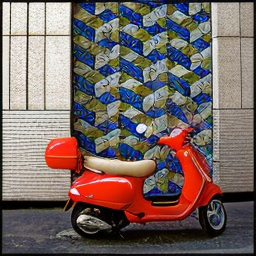} &
    \includegraphics[width=0.24\textwidth]{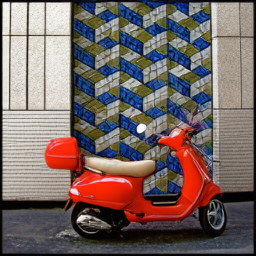} \\
    \includegraphics[width=0.24\textwidth]{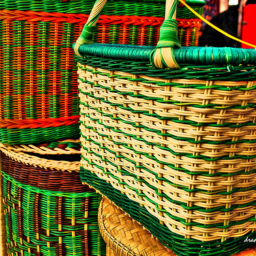} &
    \includegraphics[width=0.24\textwidth]{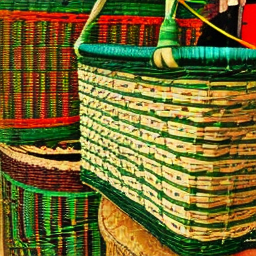} &
    \includegraphics[width=0.24\textwidth]{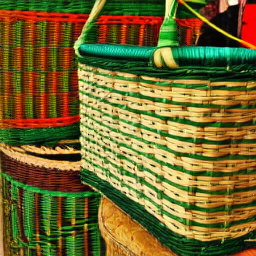} \\
    \includegraphics[width=0.24\textwidth]{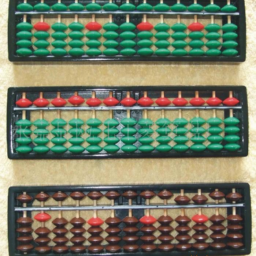} &
    \includegraphics[width=0.24\textwidth]{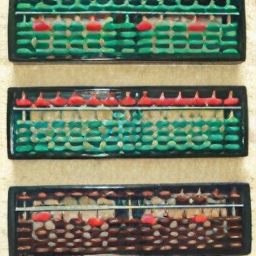} &
    \includegraphics[width=0.24\textwidth]{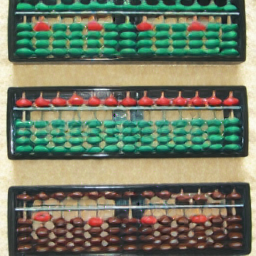} \\
  \end{tabular}
  \captionsetup{skip=3pt}
  \caption{
  \textbf{Reconstruction comparison.} Original images (left), and reconstructions from the same \sdvae latents by the \sdvaeftema decoder (middle) and by our semantic-aware decoder with \reglue semantics (right). For generated samples, see~\figureref{fig:qualitative} in the main paper.}
  \label{fig:supp_recon}
  \vspace{-10pt}
\end{figure*}

\section{Limitations}
\label{sec:supp_limitations}

We evaluate semantic-aware decoding on six diffusion- and flow-based generators. Our experiments cover several forms of semantic conditioning, with both global and patch-level targets derived from \dinovtwob. Future work could explore whether similar decoding gains hold beyond the latent diffusion models studied here (\sit, \repa, \reg, \redi, \reglue, \coredi), i.e., for other generative models trained on the autoencoder's latent space.
Further, for proxy and projected semantics, future work could compare alternative VFMs and test whether global features from one encoder complement patch features from another. Finally, it remains an open question whether the gains from semantic-aware decoding increase further when decoders are trained on larger image collections. LAION-Aesthetics, one of the datasets used to fine-tune \sdvaeftema~\citep{sdvaeftema}, offers a natural setting for exploring this question in future work.

\end{document}